# Towards geological inference with process-based and deep generative modeling, part 2: inversion of fluvial deposits and latent-space disentanglement

Guillaume Rongier[1] and Luk Peeters[2]

[1]*Department of Geoscience & Engineering, Delft University of Technology, Delft, The Netherlands*
[2]*VITO Digital Water & Soils, Mol, Belgium*

**Abstract** High costs and uncertainties make subsurface decision-making challenging, as acquiring new data is rarely scalable. Embedding geological knowledge directly into predictive models offers a valuable alternative. A joint approach enables just that: process-based models that mimic geological processes can help train generative models that make predictions more efficiently. This study explores whether a generative adversarial network (GAN) – a type of deep-learning algorithm for generative modeling – trained to produce fluvial deposits can be inverted to match well and seismic data. Four inversion approaches applied to three test samples with 4, 8, and 20 wells struggled to match these well data, especially as the well number increased or as the test sample diverged from the training data. The key bottleneck lies in the GAN's latent representation: it is entangled, so samples with similar sedimentological features are not necessarily close in the latent space. Label conditioning or latent overparameterization can partially disentangle the latent space during training, although not yet sufficiently for a successful inversion. Fine-tuning the GAN to restructure the latent space locally reduces mismatches to acceptable levels for all test cases, with and without seismic data. But this approach depends on an initial, partially successful inversion step, which influences the quality and diversity of the final samples. Overall, GANs can already handle the tasks required for their integration into geomodeling workflows. We still need to further assess their robustness, and how to best leverage them in support of geological interpretation.

## 1 Introduction

By providing raw materials and space for storage and infrastructure, the subsurface has grown into an essential resource for our society (van Ree et al., 2024). But operating in the subsurface comes with considerable costs in a limited physical space, making it essential to properly assess the economic and environmental viability of a project ahead of its implementation (e.g., Stones & Heng, 2016; Volchko et al., 2020; Lundin-Frisk et al., 2024). This cannot be achieved without characterizing the variations of the subsurface's physical properties (e.g., de Marsily et al., 2005; Simmons et al., 2023; Hinsby et al., 2024). Unfortunately the same constraints apply during exploration: acquiring subsurface data remains expensive and can be limited to certain areas because of extra constraints at the surface. And not all subsurface data are equal: they have different spatial footprints, different resolutions, and some measure the properties of interest only indirectly. To make matters worse, the subsurface is strikingly heterogeneous, and that heterogeneity is non-stationary, implying that any knowledge of the variations of a property at one location does not necessarily translate to another (Hoffimann et al., 2021).

Therefore, in most settings, we either lack data or our data are not informative enough to make precise and accurate predictions of subsurface properties. This explains the early adoption of probabilistic approaches for subsurface modeling: in a data-poor context, an estimation of uncertainty is just as important for decision making as an estimation of a property of interest. But how should we proceed when uncertainties are too large to make a decision? Acquiring more data might sound like an obvious choice, but it comes with substantial costs, and we must make sure that the newly acquired data are informative enough to actually reduce uncertainties. Taking a Bayesian perspective gives us another option: using more informative priors (Caers, 2018). The standard approaches for geological modeling rely mainly on Gaussian and marked point processes (e.g., Deutsch & Journel, 1992; Deutsch & Wang, 1996), which are statistical in nature and focus on geometries rather than geological processes. In that sense, they only encode a weak geological prior, leading to a huge potential for improvement.

Process-based models, which encode our knowledge of geological processes into numerical models, should be an obvious choice to strengthen our priors. Unfortunately, such models come with large computational costs, which only grow larger as physical processes are represented in more detail, allowing for more precise predictions. Since fitting them to some data can require thousands of model runs, those costs partly explain why they are seldom used for subsurface modeling. As shown by Rongier and Peeters (2025c), deep generative models can offer us a way forward by acting as emulators: instead of using a process-based model directly, we use it to build a training dataset for a deep-learning model, which, once trained, can generate representations of subsurface properties much faster than the process-based model. In this way, a generative adversarial network (GAN) – a type of deep generative model – can reproduce the non-stationarity and details of fluvial deposits, providing us with a stronger geological prior.

Now the remaining question is how to best use that prior to make predictions based on subsurface data. This general process of matching the realizations from a generative model with some data is known as conditioning in geostatistics and geological modeling. For GANs, the deep-learning community uses a more specific terminology:

- Conditioning implies that the data-matching process is part of the generative process (e.g., Song, Mukerji, Hou, et al., 2022). So the GAN's architecture is adapted to take some data as inputs and data matching is an integral part of the training process. This makes predicting more straightforward, since samples directly match the data when generated, but less flexible, since a conditional GAN always requires the specific data types it was trained on as inputs.

- Inversion implies that the data-matching process is separate from the generative process (e.g., Dupont et al., 2018; Laloy et al., 2018). So the GAN's architecture is unconditional and has to be pretrained; then the data-matching process uses optimization or sampling to select the samples matching the data from the GAN's latent space, i.e., its input parameters. This makes predicting more flexible, since any kind of data can be matched without re-training the GAN itself, but less practical, since it involves an extra step that may require some tuning.

In this article, we build upon the work of Rongier and Peeters (2025c) to test the applicability, benefits, and limitations of GAN inversion when predicting some properties of fluvial deposits. We focus more specifically on three main questions:

1. How do GAN inversion approaches perform on continuous, non-stationary 3D properties?
2. How do these approaches perform when having to integrate an increasing number of (diverse) data?
3. How do these approaches perform with predicting extra properties for which data are lacking?

## 2 Materials and methods

This section provides a general overview of the approaches used in this work. We refer readers who are after more details to the articles cited in the subsections below and to the implementation that supports our study (Rongier & Peeters, 2025b).

### 2.1 Generative adversarial networks

A generative adversarial network is a type of deep generative model trained through an adversarial scheme (Goodfellow et al., 2014): a deep neural network called the generator learns to generate new samples, while another network called the discriminator learns to determine whether a sample comes from the training data or from the generator. The generation process starts from a vector of random numbers, which represents the input space of the generator, called the latent space. During training, the discriminator is confronted with both kinds of samples, while the generator never sees a training sample. Instead, it gets the assessment of the discriminator on its samples, from which it must learn to fool the discriminator. This explains the term "adversarial": both networks have competing objectives leading to a zero-sum game. And it makes GANs harder to train than other deep-learning models, because if one network overcomes the other, both networks stop learning.

Many techniques have been developed to stabilize GANs' training and scale it up to larger samples (e.g., Heusel et al., 2017; Miyato et al., 2018; Brock et al., 2019). Rongier and Peeters (2025c) have shown that those techniques are already enough for a GAN to consistently learn to generate fluvial deposits as simulated by a process-based model. In their specific case, it was even possible to simplify the GAN architecture while preserving the samples' quality. Here, we use their architecture 4 with a latent size of 128 (see their figure 6), which starts from a version of DCGAN – a simple convolutional GAN (Radford et al., 2016) – adapted for 3D generation and:

- Replaces the ReLU activation functions in the generator by Leaky ReLU functions.

- Replaces the binary cross entropy in the loss function and the sigmoid function as last activation in the discriminator by a binary cross entropy with logits and a linear function.

- Replaces the $\beta_1$ and $\beta_2$ of 0.5 and 0.999 in the Adam optimizer with 0 and 0.99.

- Adds spectral normalization in the generator and discriminator.

- Replaces the convolutional layers in the generator and discriminator with residual blocks.

- Removes the batch normalization in the discriminator.

- Adds a $R_1$ regularization to the discriminator.

All those changes are borrowed from a more advanced GAN, called BigGAN (Radford et al., 2016).

A trained GAN allows us to quickly generate a sample from a latent vector, but not the other way around: given a sample, we cannot directly retrieve its position in the latent space. Instead, we have to rely on inversion techniques. Here, we focus on four approaches that have already been successfully applied to geological cases:

- Latent optimization is the simplest approach, in which an initial latent vector is randomly selected, then optimized using a gradient-based approach to minimize the difference with the data (Dupont et al., 2018). This optimization step has to be repeated for each sample to invert.

- An inference network is a fully-connected deep neural network that learns a new latent space which focuses on generating samples that match the data (Chan & Elsheikh, 2019). Once this network is trained, any number of samples can be quickly generated.

- Variational inference uses a deep neural network – here based on normalizing flows – to approximate the posterior distribution through optimization (Levy et al., 2023). It is an alternative to Monte Carlo sampling methods, and it shares similarities with the inference network.
- Markov chain Monte Carlo (MCMC) inversion provides an approximation to the posterior distribution through sampling with, in the case of the DREAM$_{(ZS)}$ algorithm used here, multiple chains with possible jumps between them to avoid aberrant trajectories (Laloy et al., 2018). It provides a more robust but more expensive Bayesian perspective on inversion.

In all those approaches, the generator's weights are kept constant during inversion.

## 2.2 Training and testing data

Our training and testing data rely on FluvDepoSet (Rongier & Peeters, 2025a), which contains 20 200 synthetic 3D realizations of fluvial deposits, each simulated using the Channel-Hillslope Integrated Landscape Development Model (CHILD) (Tucker, Lancaster, Gasparini, Bras, & Rybarczyk, 2001; Tucker, Lancaster, Gasparini, & Bras, 2001). To assess the reproducibility of the inversion results, we built three GAN models, which all have the same architecture and hyperparameter values but are initialized with different random noise values. Each model was trained using the first 20 000 realizations only, following the procedure detailed in Rongier and Peeters (2025c): each realization is cropped and filled to keep only the fluvial deposits, then randomly cropped to extract a training sample with $128 \times 128 \times 16$ cells of size $50 \times 50 \times 0.5$ m that contains two properties – the fraction of coarse sediments and the normalized deposition time.

The realizations 20 001 to 20 200 were set aside for testing. We extracted a test sample with $128 \times 128 \times 16$ cells from the middle of the fluvial deposits in each realization – starting at index (0, 36, 10). To keep our study computationally tractable, only the first three test samples – 20 001, 20 002, and 20 003, hereafter called test sample 1, 2, and 3 – are used as ground truths from which we extracted data for inversion (figure 1). The other test samples are used to assess the quality and diversity of the inverted samples.

Our data-generating process aims at mimicking the development phase of a subsurface project, with an inversion focusing on subsurface characterization. As such our case study does not target a specific application; we just assume that the coarse sediments represent the target of interest. Data about the distribution of those coarse sediments can be acquired in situ by collecting cores and downhole geophysical measurements from wells – leading to high-resolution data with a limited spatial coverage – or remotely by collecting geophysical measurements from the surface – leading to low-resolution data with a broad spatial coverage.

Here, in situ data of coarse-sediment fraction were directly extracted from test samples 1, 2, and 3 along 20 vertical wells. The locations of those wells were selected sequentially and randomly, with an exclusion zone to avoid that wells end up implausibly close (appendix A). The first 4 wells are legacy wells that were targeting different intervals, while the last 16 wells target these specific coarse deposits. We then defined three well datasets for each test sample: one with the 4 legacy wells, one with the 4 legacy wells and 4 extra wells, and one with the 4 legacy wells and the 16 extra wells (figure 1). Our goal is to assess how each inversion technique performs as the number of data increases. Each well only contains the fraction of coarse sediments, and another goal is to check whether we can extract any valuable information on deposition time from the inversion. We stopped at 20 wells because this would already be a high well density in a real setting, and one can question the usefulness of a strong geological prior in settings with higher densities.

To assess how some inversion techniques perform when exposed to different data types, we also generated geophysical data acquired remotely from the surface, more specifically 3D seismic data. Seismic data are acquired by emitting acoustic waves that reflect off layers with contrasts in density or seismic velocity. The recorded reflections can be processed into a 3D cube showing some subsurface structures. Our fluvial deposits are made of a mixture of unconsolidated sediments, with a coarse and a fine fraction. We considered that they remained unconsolidated after burial and computed density and seismic velocity from the constant-clay model for shaly sands combined with Gassmann's equation (Avseth et al., 2005) (appendix A). A 3D seismic cube was then generated for each of test samples using a 3D convolution approach based on a point-spread function (Lecomte, 2008; Lecomte & Kaschwich, 2008; Lecomte et al., 2015). Our samples are thin compared to the vertical resolution of seismic data, so we added a structureless over- and underburden, for a total of 18 extra meters vertically (figure 1).

## 2.3 Validation

To validate the inversion itself, we use the mean absolute error $e$ between the true property $y$ and the inverted property $\hat{y}$ at a given location $i$ for a set of locations $n$:

$$e = \frac{1}{n} \sum_{i=1}^{n} |y_i - \hat{y}_i| \qquad (1)$$

We then define two errors depending on the set used: the inversion error is the error with the well data, while the generalization error is the error with the entire ground truth. Since the well data only contain the fraction of coarse sediments, the inversion error is limited to that property. The generalization error however can be computed for the fraction of coarse sediments and for the normalized deposition time. This allows us to assess whether we can get any insight on deposition time and history from fraction data alone.

Matching continuous properties always leads to some residual error. In a real case study, where priors are an approximation of reality at best and misspecified at worst, we consider that an inversion error of 10 % on the fraction of coarse sediments would already be useful to support downstream tasks. However, here, our prior is based on the exact same model used to simulate the test samples. So there is no misspecification, only a slight approximation from using a GAN instead of CHILD, the original landscape evolution model. In that context, we consider that an inversion error of 1 % on the fraction of coarse sediments is the maximum we should tolerate. As an initial test, we randomly generated 300 samples to visualize the prior defined by each of the three GAN models (figure 2, top), and repeated the operation for each case study to assess the errors before inversion (figure 2, bottom). In most cases, the error with the well data (i.e., the inversion error) is above 10 %.

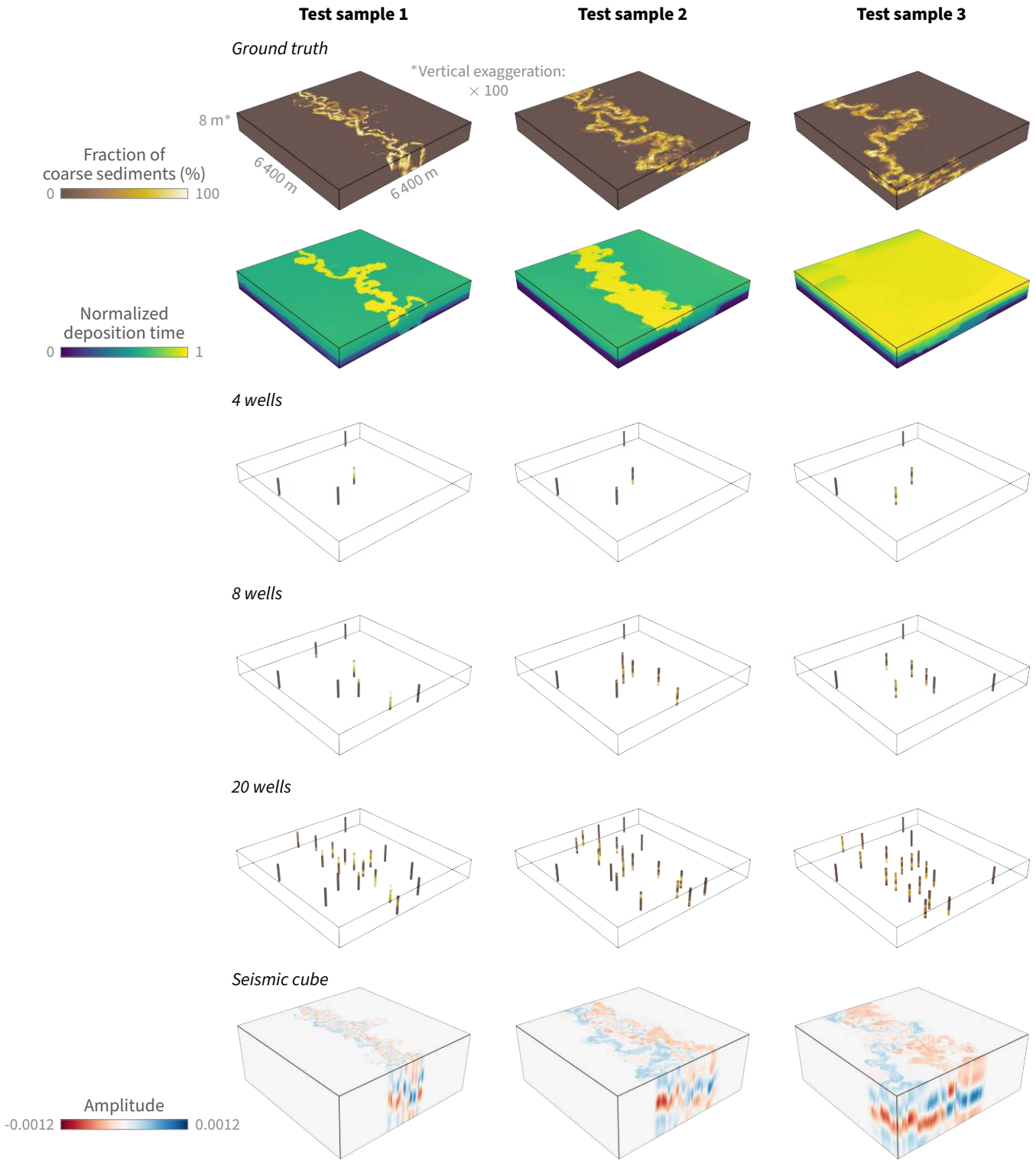

**Figure 1** Test samples and data extracted from those samples.

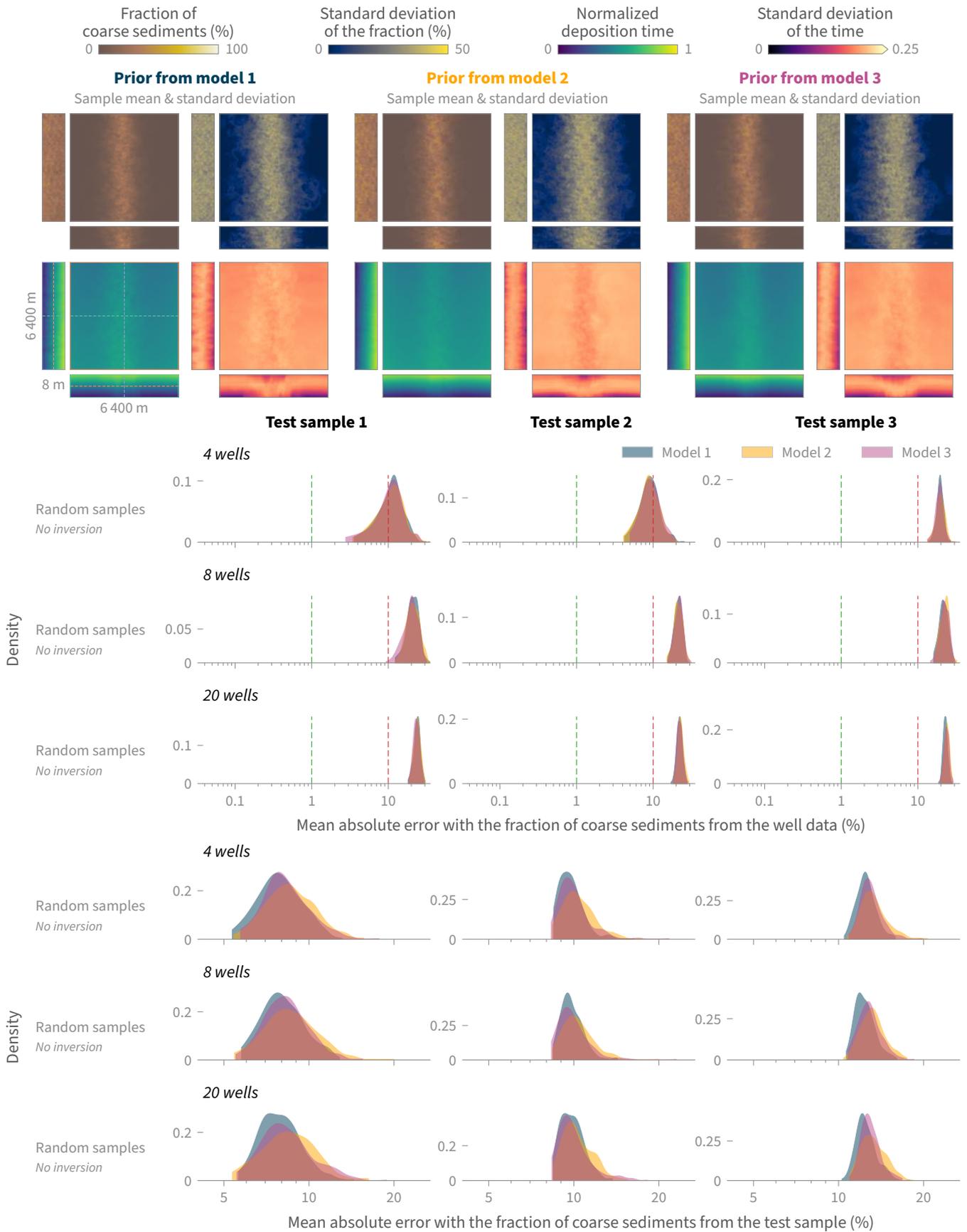

**Figure 2** Mean and standard deviation of the fraction of coarse sediments and of the normalized deposition time for 300 random samples from the three GAN models, which show us the prior defined by each model; error between 300 random GAN samples and the well data, and error between the same samples and the test samples, which show us the mismatch between the prior and the data and ground truth (i.e., inversion and generalization errors). The slices go through the center of the samples, as shown on the normalized deposition time of model 1.

To further analyze the quality and diversity of the inverted samples, we follow the approach laid out by (Karras et al., 2018) and (Song et al., 2021): measuring the sliced Wasserstein distance averaged over multiple levels of a Laplacian pyramid between patches extracted from inverted samples and from the 200 test samples, then mapping those distances into a 2D space for visualization and analysis using multidimensional scaling (MDS).

## 3 Results

All the results presented in this section were computed using either one or two NVIDIA A100 80GB GPUs. All our experiments are based on generating and inverting 300 samples using the GAN's generator.

### 3.1 Initial GAN inversions

We start with only the well data, and focus on the inversion error, which shows distinct behaviors depending on the test sample, the number of wells, and the inversion approach (figure 3). The latent optimization is the only approach that fails systematically to pass the 1 % threshold (see figure B.1 for the training losses). The other approaches pass that same threshold on the test sample 1 with 4 wells, and the 10 % threshold on most other cases. With 8 and 20 wells, the inference network and the variational inference show signs of collapse, with all inverted samples being similar or even identical (see figures B.2 and B.3 for the training losses). While less obvious thanks to the different chains, samples from the MCMC follows a similar fate, and the convergence diagnostic $\hat{R}$ is never satisfied for all the latent parameters (see figure B.4 for the errors during sampling).

Overall, inversion errors clearly increase from test sample 1 to 2 to 3, and from 4 to 8 to 20 wells. This observation can already be made on samples randomly drawn from the latent space, without inversion (figure 2). Increasing the number of wells increases the constraints on the inversion, so higher inversion errors are to be expected. The increase of error between test samples is consistent with the distribution of samples in FluvDepoSet (figure 2 and the samples shown in Rongier & Peeters, 2025c), which mostly have a centered, relatively narrow channel belt. Test sample 1 is the closest to that dominant mode. Test sample 2 shows a larger channel belt, which must be less represented in the latent space, so more difficult to find. Test sample 3 shows a channel belt that diverges from the center, which is rare in the training samples, making it even less represented and much harder to find.

Visualizing the structure of the GAN's latent space is challenging because of its 128 dimensions. To get a glimpse through the variations of inversion error, we turned to a technique used to visualize the loss landscape of neural networks (Li et al., 2018). Following Liu et al. (2023), we used the latent vectors obtained by the latent optimization and principal component analysis (PCA) to find two principal directions. From those two directions, we can define a 2D slice through the full error landscape. That slice is centered at the mean latent vector, and expand from -20 to 20 along the two directions (figure 4 for model 1, B.5 for model 2, and B.6 for model 3). While it only gives us a limited view of the true error landscape, it comforts our initial observations. The minimal error tends to increase with the number of wells and from test sample 1 to 2 to 3. And the error landscape becomes more rugged as the number of wells increases, with what appears like more local minima, or at least more local variability in the error. This might confirm Rongier and Peeters (2025c)'s suggestion that the latent space is entangled, so that samples with similar characteristics are located in different parts of the latent space. It would make inversion more difficult, especially if those different parts are very localized. The error landscape of test sample 2 for 4 wells also show a large, seemingly flat area, which would be problematic for gradient-based optimization if it translates to the full latent space.

Training multiple GAN models led to little variability in terms of stability and sample quality (see Rongier & Peeters, 2025c, figure 1). However, the inversion leads to more variability between models, with in general model 1 performing slightly better, model 3 slightly worse. And while the error landscapes for the three models show similar general characteristics, they are locally completely different. All this suggests that the latent structure can change significantly depending on the GAN's initialization.

### 3.2 Approaches for latent restructuring

All the inversion techniques used in the previous section have been successfully tested on direct data, but mainly on stationary discrete properties in 2D with two facies, channel and floodplain. Switching to a non-stationary continuous property in 3D leads to jump in complexity. Yet, our assumption going forward is that the main bottleneck arises not so much from the inversion approaches themselves, but from an entangled latent space.

We tested that assumption by using four approaches to restructure the latent combined with the simplest inversion technique, the latent optimization:

- Bigger GAN uses architecture 8 of Rongier and Peeters (2025c) instead of 4 (see their figure 1), so an architecture closer to the full BigGAN. The three key changes compared to architecture 4 are doubling the residual blocks, so doubling the number of weights that can learn, changing the random initialization of the weights in the convolutional layers, and adding a skip connection from the latent vector to the residual blocks.

- The latent-size-512 model uses a latent size of 512 instead of 128 with architecture 4. Overparametrizing the latent space has been shown to improve inversion with a different GAN architecture (Poirier-Ginter et al., 2022).

- GAN conditioning implements the conditioning strategy of BigGAN in architecture 4 (Brock et al., 2019) to five of the seven parameters that vary between the samples of FluvDepoSet: the coarse grain diameter, the fine grain diameter, the bank erodibility, the average of the river aggradation rate through time, and the average of the mean storm rainfall through time (figure C.3). This gives us five extra parameters during inversion that are geologically consistent.

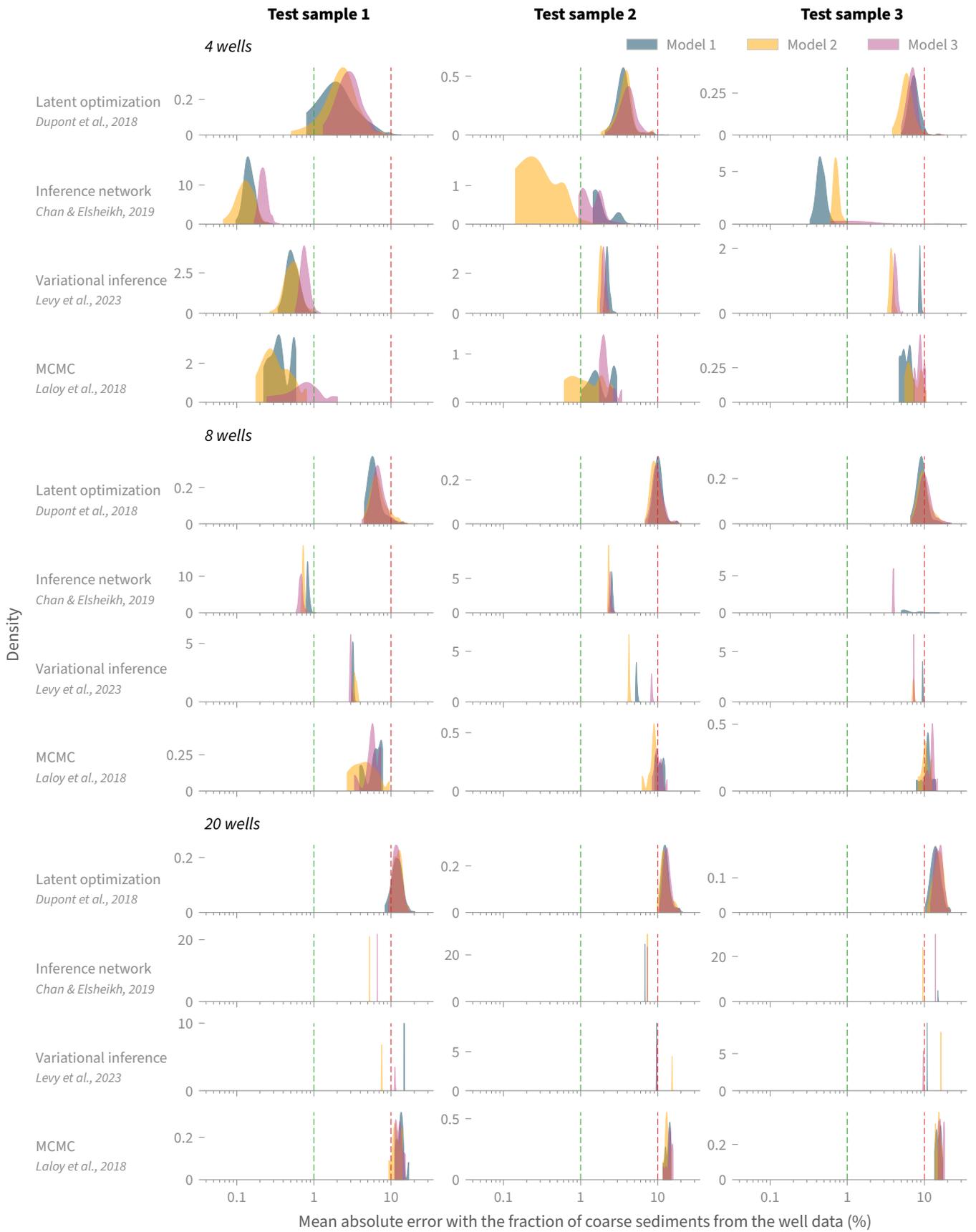

**Figure 3** Inversion error between 300 inverted GAN samples and the well data for four inversion approaches.

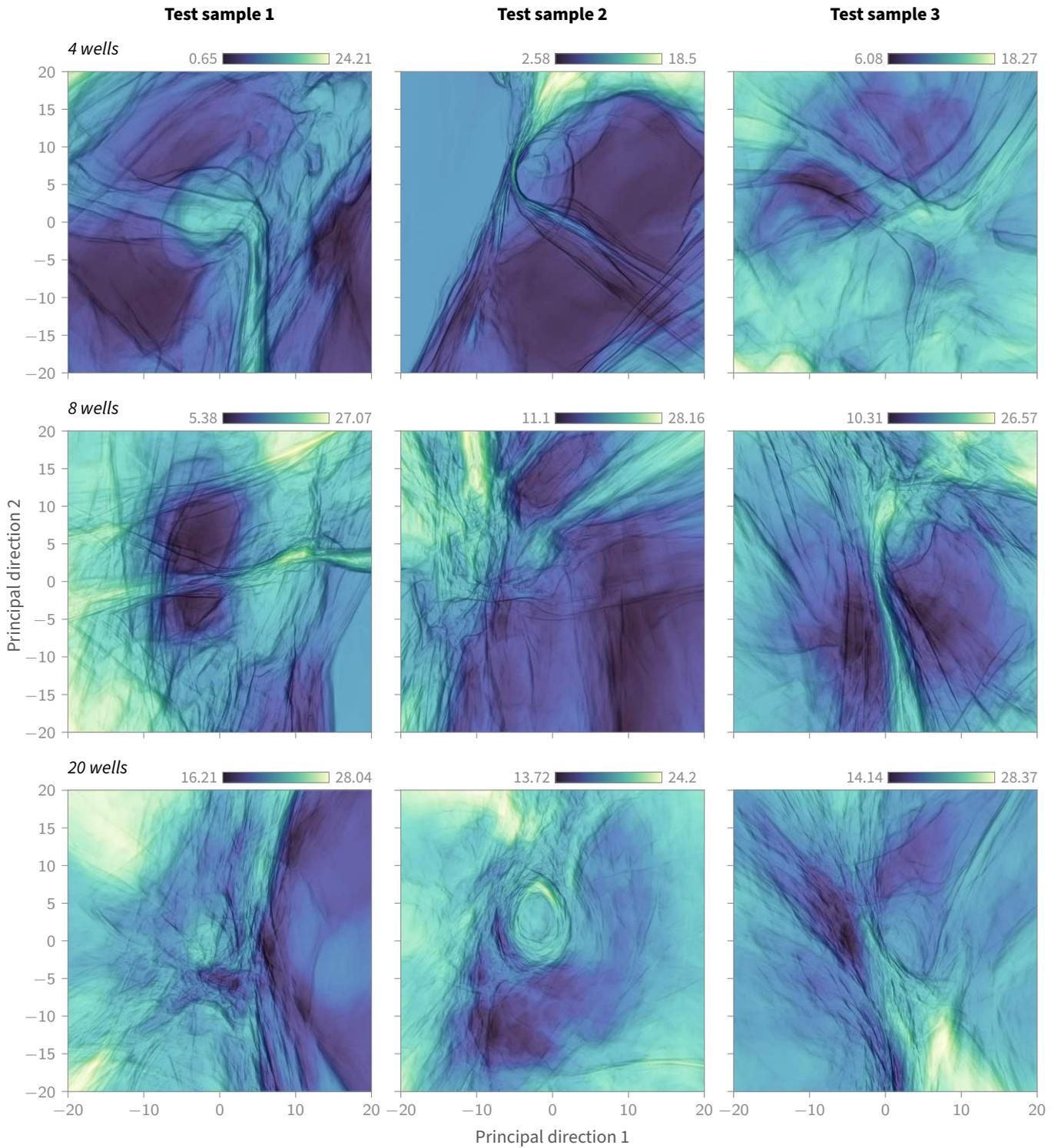

**Figure 4**  Inversion error landscape for model 1.

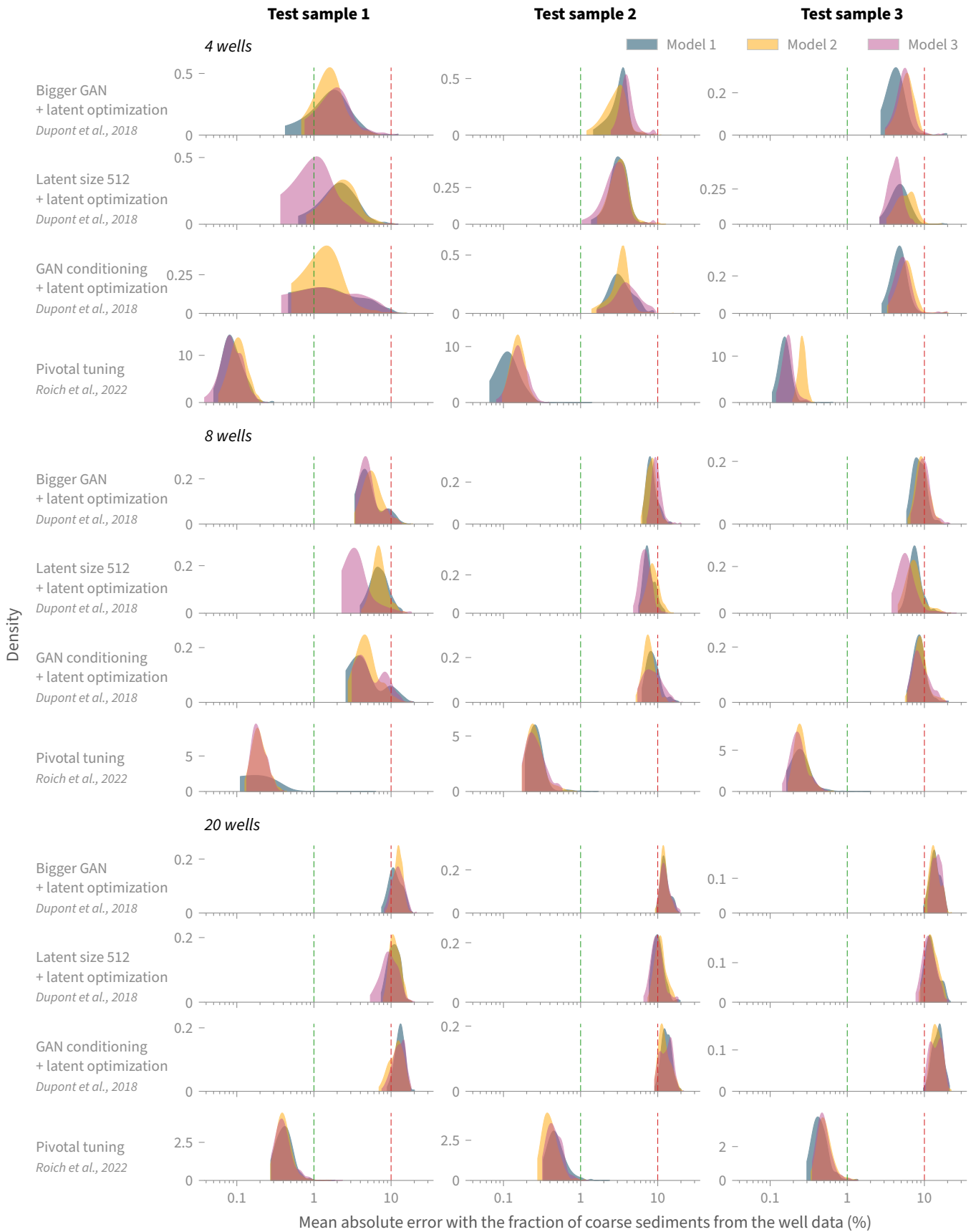

**Figure 5** Inversion error between 300 inverted GAN samples and the well data for four latent restructuring approaches.

- Pivotal tuning pushes restructuring further by fine-tuning the latent space to better align with the data (Roich et al., 2022). Contrary to the previous approaches, it uses the latent vectors obtained by latent optimization as input and updates the weights of the generator. Here, we use the discriminator in the perceptual loss instead of the original VGG network, which is only defined for 2D images.

All those approaches decrease the errors for all cases, with the most dramatic improvement coming from the pivotal tuning (figure 5). Using a bigger GAN leads to more modest improvements, although the behavior of the different models seems more consistent (see figure C.1 for the training losses). Using a larger latent and conditioning shows similar improvements, which vary between different models (see figures C.2 and C.4 for the training losses). The pivotal tuning is the only technique that passed the $1\,\%$ threshold on the vast majority of its 300 samples for all cases (see figure C.5 for the training losses). All this suggests that restructuring the latent improves the inversion, and that being able to fine-tune the generator is a huge advantage.

### 3.3 Data integration and generalization

Of course, matching the data is only part of what we are interested in: ultimately, we want to make more accurate predictions, including of uncertainty. So we need to look at the generalization error (figure 6). From now on, we focus on pivotal tuning, because it was the only approach capable of consistently reaching the $1\,\%$ threshold, and on latent optimization, because pivotal tuning builds upon it. First, the generalization error decreases for all test samples when going from 4 to 20 wells, showing that the inversion can exploit the extra data. Second, the generalization error increases from test sample 1 to 2 to 3, which is already visible with purely random samples (figure 2). This confirms that the prior encoded in the GAN is closer to test sample 1 than test sample 3, explaining the difficulty in inverting that test sample. Third, there is little to no difference in generalization error between the latent optimization and the pivotal tuning with only the well data, although there is a huge difference in inversion error (figure 5). This suggests that either the pivotal tuning only makes changes at the data locations, or that well data provide little information away from them. On a similar note, the latent optimization systematically decreases the inversion error compared to fully random samples (figures 5 and 2), yet it increases the generalization error with 4 and 8 wells. In principle, this is encouraging: if we want to properly quantify the impact of uncertainties, we need to capture all the distributions of deposits that match the data, including those far from the ground truth. Again, the three GAN models tend to behave differently.

Adding the seismic data has a huge impact on the generalization error (figure 6, see figures D.1 and D.2 for the training losses, and figure D.3 for the inversion errors). It systemically decreases after latent optimization, a decrease that is quite consistent across models. But it also keeps decreasing after pivotal tuning, contrary to the case without seismic data. This confirms that the latent optimization is not capable of fully exploiting the data, and that being able to fine-tune the generator – so to alter the latent space – is essential.

To push our analysis further, we can look at the mean fraction of coarse sediments and its standard deviation across inverted samples for each test sample, number of wells, and GAN model (figures 7, 8 and 9 for model 1, E.1, E.2 and E.3 for model 2, and E.4, E.5 and E.6 for model 3). And the same observation stands: the more data we add, the closer we get to the ground truth, no matter the case and the model. The seismic data plays a key role in this, which is in line with its large spatial coverage. The pivotal tuning with 20 wells and the seismic cube even captures the general structure within the channel belt on the mean fraction (e.g., figures 7). With fewer data, it is also capable of capturing multiple scenarios for the location of the channel belt (e.g., figures E.3). Differences between the latent optimization and the pivotal tuning are often localized, with more global changes on the channel belt observed when adding seismic data or with test sample 3 (e.g., figures 9). In two cases, unrealistic checkboard patterns appear in the channel belt after pivotal tuning (e.g., figures E.3, 20 wells). One key difference between latent optimization and pivotal tuning comes from the reduction in uncertainty from well data: latent optimization leads to large exclusion zones around data with finer sediments, while pivotal tuning leads to much more local reductions of uncertainty (e.g., figures 7, 20 wells). Channels can be quite narrow, with abrupt changes in direction due to meandering (as visible in test sample 1), so in such setting knowing that we have coarser or finer sediments at a given location tells us little about what to expect around that location. From that perspective, latent optimization underestimates uncertainty, probably because it struggles to explore the latent space. Pivotal tuning on the other hand should come closer to the true uncertainty. Again, the three GAN models lead to different mean predictions and uncertainty quantification. This tends to confirm the instability of the latent representation learned during training, although we would need more inverted samples to further support that claim.

The generalization error on the normalized deposition time is less conclusive (figure 10). The decrease in error as the number of wells increases remains limited, and more focused on the upper tail of the distribution. Adding seismic data has a limited effect as well. Focusing on the cases with the most data, some incisions start to be visible, but only inconsistently between GAN models (figure 11 for model 1, figure E.7 for model 2, and figure E.8 for model 3). Similarly, uncertainties differ significantly between models, sometimes focusing on specific time intervals (e.g., figure E.8, test sample 3), sometimes showing no valuable information (e.g., figure E.7, test sample 1). These results remain too rough and inconsistent to be of much practical value, but they show some potential for extracting extra geological insights from geological modeling.

### 3.4 Sample analysis after pivotal tuning

If pivotal tuning leads to the most consistent results, we still need to analyze its effect on the samples themselves. For this, we can compare two extreme cases with the test samples: the case with only 4 wells, which has the smallest constraint on pivotal tuning, and the case with 20 wells and the seismic cube, which has the biggest constraint on pivotal tuning (figures 12, 13 and 14 for model 1, E.9, E.10 and E.11 for model 2, and E.12, E.13 and E.14 for model 3). Since the normalized deposition time showed less improvement and is less valuable for downstream tasks, we only use the coarse fraction of sediments for this comparison. On all the multidimensional scaling maps, the inverted samples from 20 wells and the seismic cube are closer to the test samples than those from 4 wells, and closer together, which is what can be expected. The diversity of patterns on the 4-well case is almost as large as that of the test samples, which is also to be expected considering the small number of constraints. However, that case often ends up along the margin of the test samples, with little (figure 13) to no overlap (figure E.14).

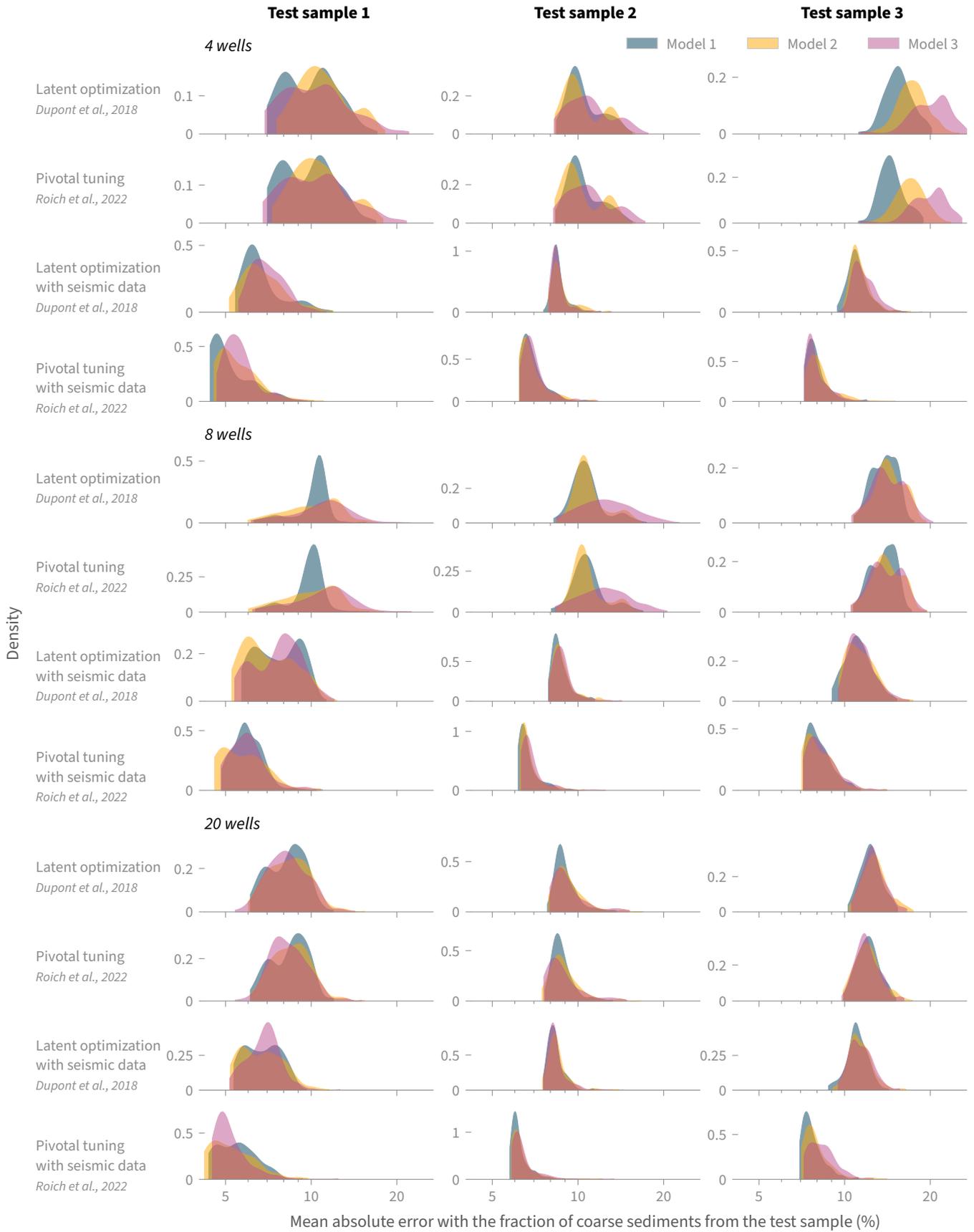

**Figure 6** Generalization error between 300 inverted GAN samples and the test samples for two inversion approaches without and with seismic data.

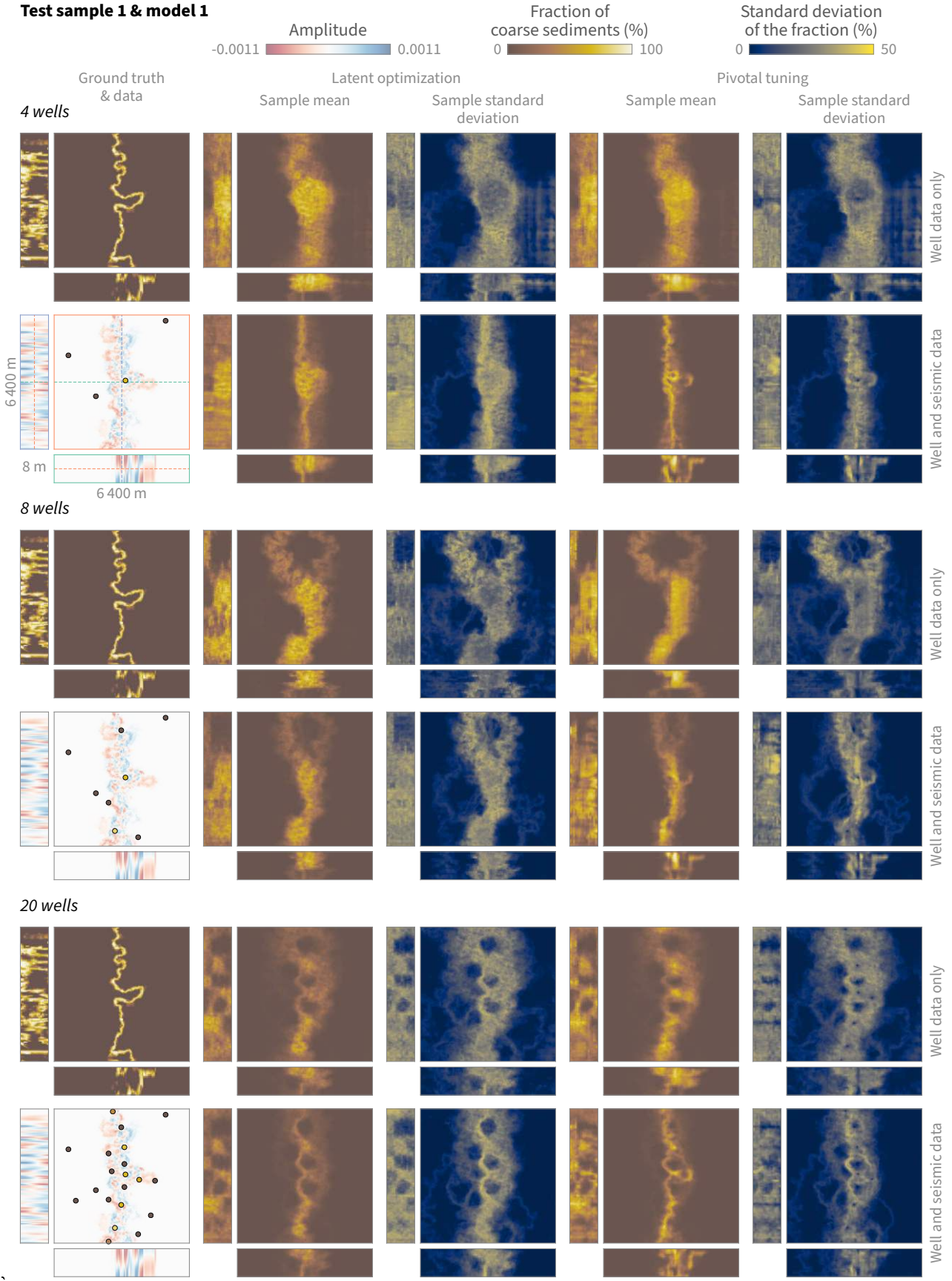

**Figure 7** Comparison between the mean and standard deviation of the fraction of coarse sediments for 300 samples inverted from the GAN model 1 with test sample 1 for two inversion approaches without and with seismic data. The slices go through the center of the samples, as shown on the data with 4 wells.

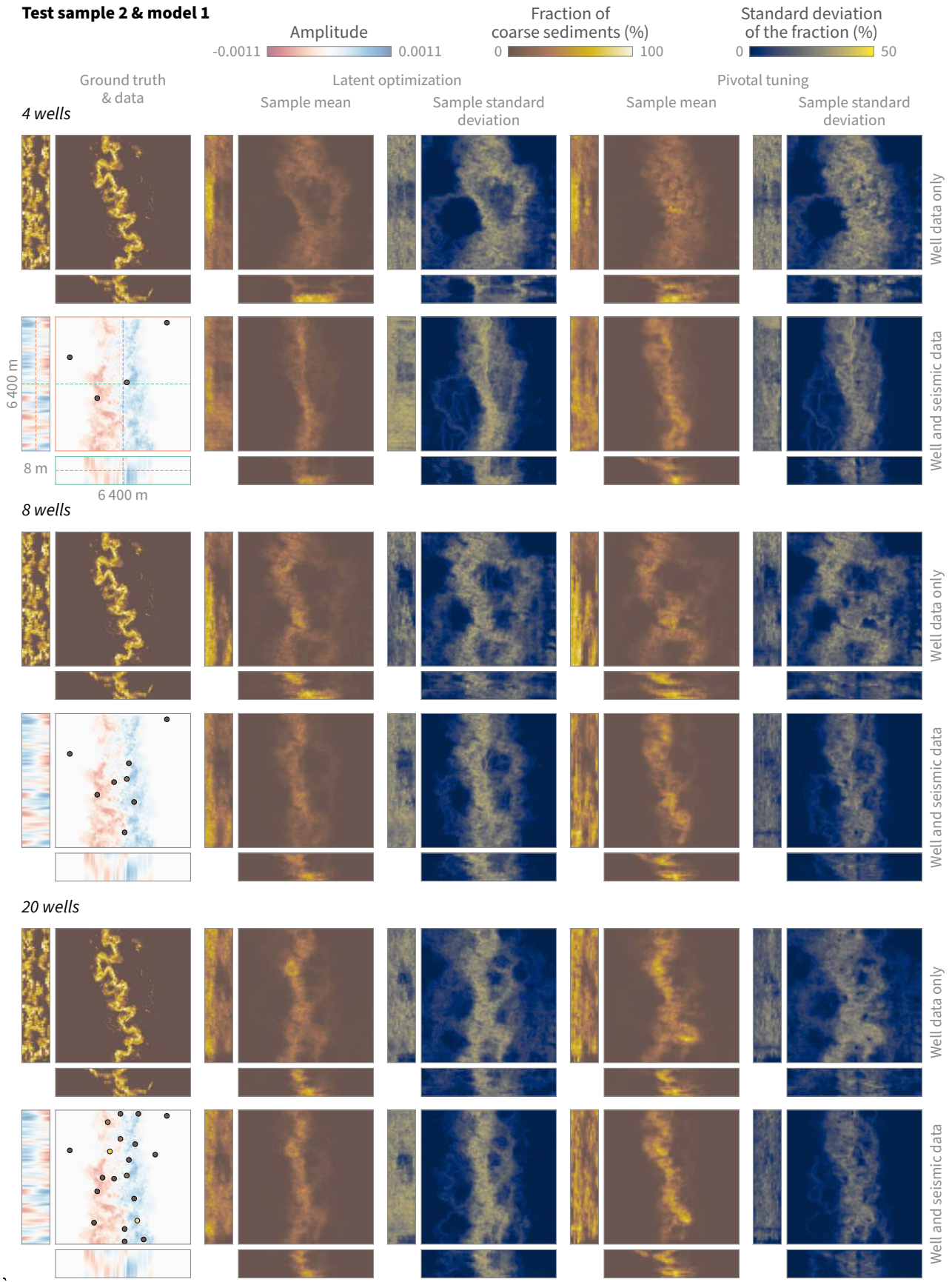

**Figure 8** Comparison between the mean and standard deviation of the fraction of coarse sediments for 300 samples inverted from the GAN model 1 with test sample 2 for two inversion approaches without and with seismic data. The slices go through the center of the samples, as shown on the data with 4 wells.

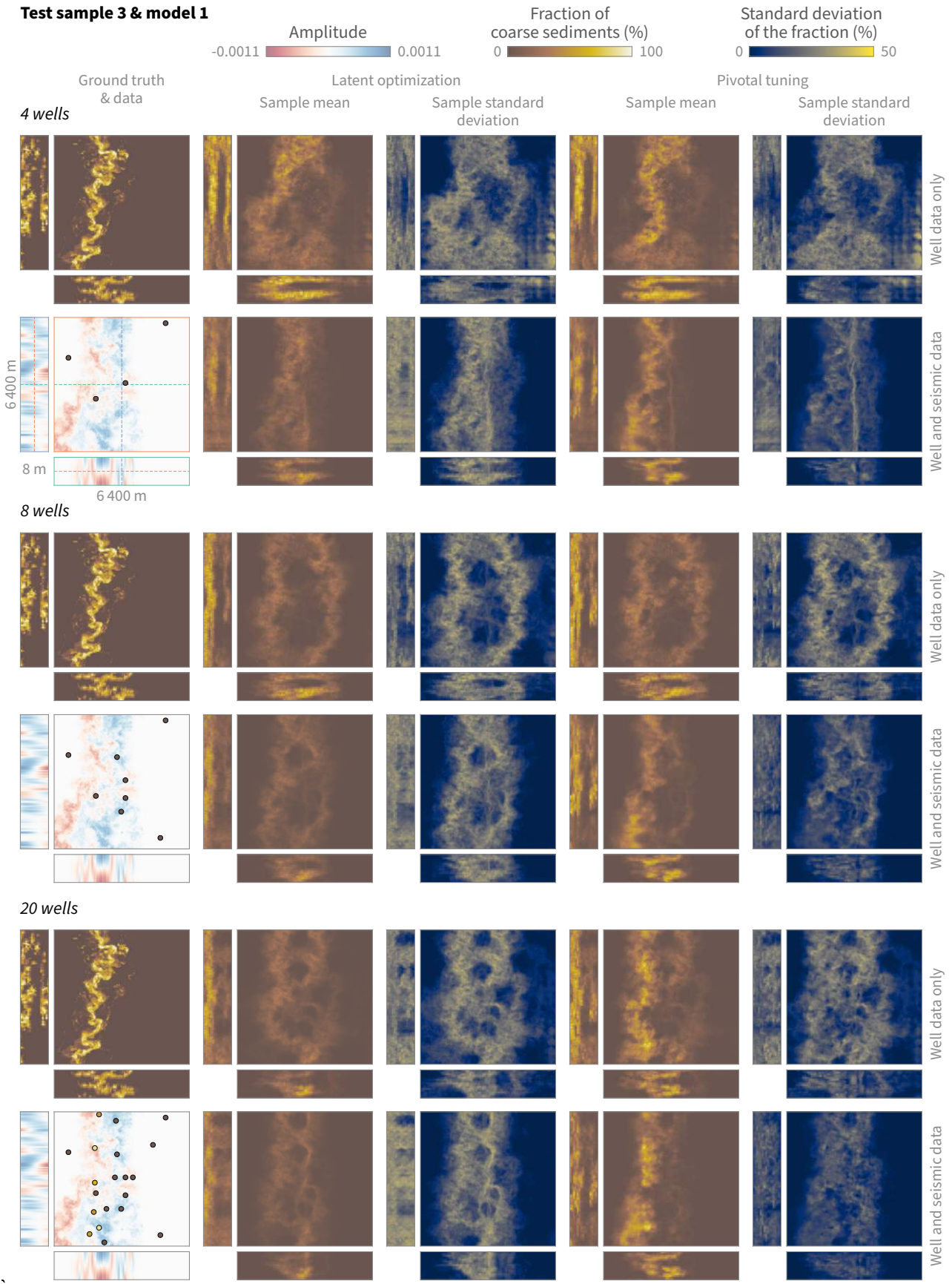

**Figure 9** Comparison between the mean and standard deviation of the fraction of coarse sediments for 300 samples inverted from the GAN model 1 with test sample 3 for two inversion approaches without and with seismic data. The slices go through the center of the samples, as shown on the data with 4 wells.

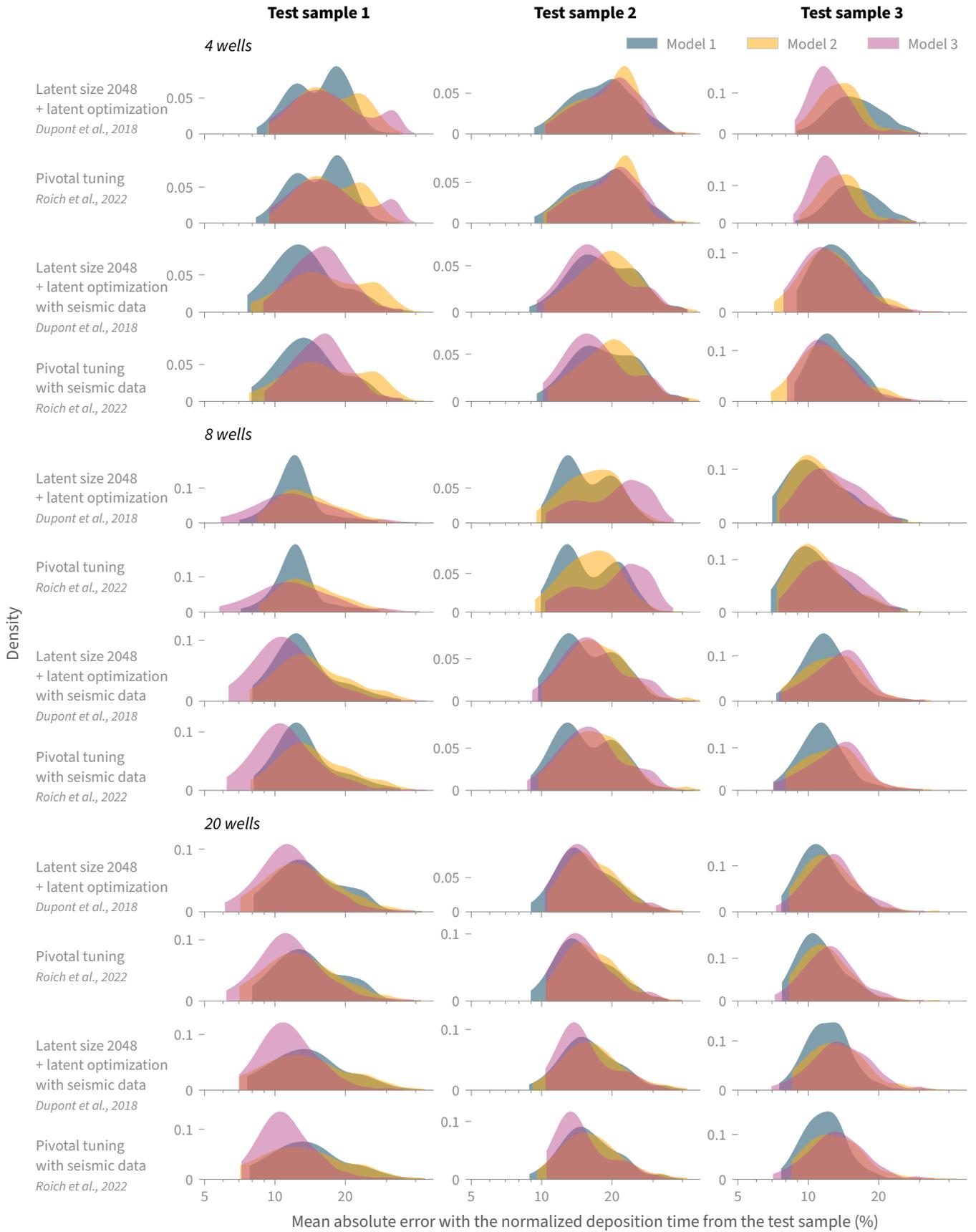

**Figure 10** Generalization error for the normalized deposition time between 300 inverted GAN samples and the test samples for two inversion approaches without and with seismic data.

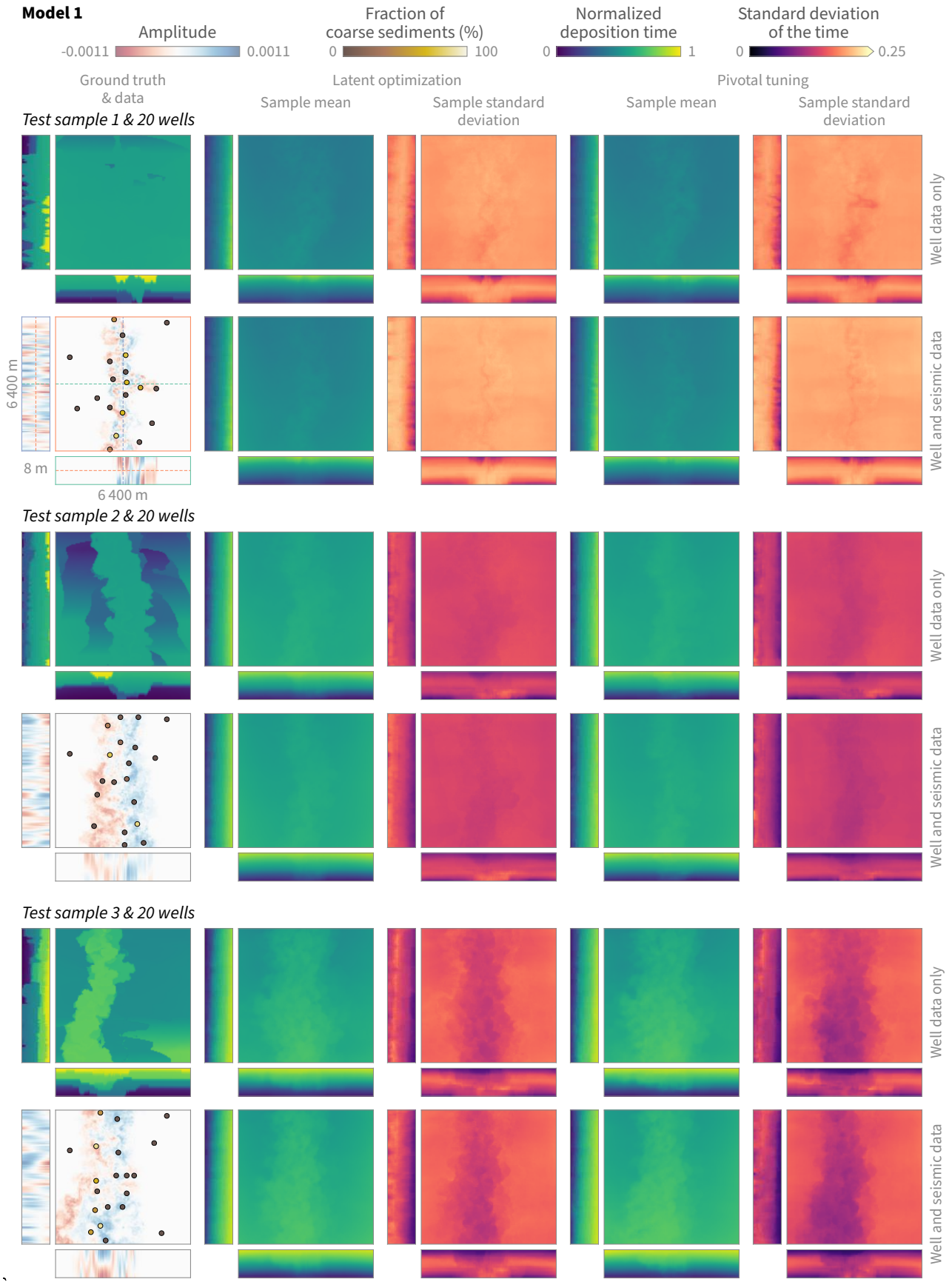

**Figure 11** Comparison between the mean and standard deviation of the normalized deposition time for 300 samples inverted from the GAN model 1 with 20 wells for two inversion approaches without and with seismic data. The slices go through the center of the samples, as shown on the data with 4 wells.

The inversion can refocus the samples to specific patterns not well represented in the test samples, so lack of overlap is not necessarily a sign of poor quality. A visual inspection of some randomly selected samples shows three main outcomes from the inversion (figures 12, 13 and 14 for model 1, E.9, E.10 and E.11 for model 2, and E.12, E.13 and E.14 for model 3). In the first outcome, pivotal tuning appears perfectly capable of making significant changes to the samples while preserving their geological plausibility (e.g., figure 14, 4 wells, and figure 12, 20 wells & seismic cube). With enough data, some samples even show features close to the ground truth (e.g., figure E.9, 20 wells & seismic cube). In the second outcome, pivotal tuning struggles to preserve geological plausibility, most likely because the samples after latent optimization are still too far from the data. This is most obvious on test samples 3 with 20 wells and the seismic cube (e.g., figures 14 and E.14). The seismic and well data are weighted to contribute equally to the loss, but this might not be the best choice, especially for latent optimization. In the third outcome, samples after latent optimization show a lack of diversity or implausible features, such as a completely disconnected channel belt or check-board patterns (e.g., figures 12 and E.14, 4 wells). These artifacts are rare when sampling randomly from the generator (see the samples visible in Rongier & Peeters, 2025c), but can become more prevalent after inversion. This might suggest that sample quality is variable across the latent space, so that sample quality should be analyzed in relation to the structure of the latent space. All in all, pivotal tuning offers more a local correction than a complete reorganization of the latent space, and latent optimization retains a huge impact on the general structure of the samples and on their quality.

## 4 Discussion

This work is an attempt to study GAN inversion as comprehensively as possible: we used multiple test samples, multiple amounts and types of data, multiple inversion techniques, and multiple pretrained GAN models. But it came with a huge computational cost, and we had to make choices to keep it manageable. As such, we used the different inversion techniques *off the shelf*, i.e., we kept the hyperparameter values used in their original studies. Just like in GAN training, we believe that tuning must remain minimal if we want deep generative modeling to be used in practice, because not all users will be nor should be experts of each approach. We still tested different hyperparameter values for each technique (e.g., weight in the loss and radius for the latent optimization, number of layers and types of activation functions for the inference network, type of neural network and number of layers for the variational inference, number of chains and scaling factor jump rate for the MCMC), focusing on the most difficult cases, those with 20 wells. While non-exhaustive, none of those tests led to a consistent and significant decrease of the inversion error. And our results are consistent with other studies. Laloy et al. (2018) have already shown on a stationary discrete property in 2D with two facies that latent optimization fails to properly match some ground-penetrating-radar data; Bhavsar et al. (2024) have shown on a stationary discrete property in 3D with multiple fluvial facies that the inference network fails to match all the data from 4 vertical wells and that its samples lack diversity. Nevertheless, our study should be taken as a base scenario with some room for improvement.

Ultimately, the key bottleneck stems less from the inversion techniques than from the structure of the latent space itself. This aspect has already been a strong point of focus in the deep-learning community (Xia et al., 2023). But what would be a proper structure for subsurface modeling remains to be seen. Is being disentangled good enough, even if it means that each latent parameter only updates samples locally? Would controlling more global, descriptive features like the width of the channel belt or the overall fraction of coarse sediments help? Would learning features related to the geological processes like the migration or the aggradation rates improve the inversion further? Inverting based on a GAN conditioned to parameters of the process-based model gives us an initial indication that it might. Deepening that answer requires a more thorough study of the conditioning quality. Some parameters lead to the expected changes in the samples, e.g., a lower aggradation rate leads to more sediment reworking and incision phases, while a higher bank erodibility leads to a faster lateral migration without incision phases (figure C.3). Others have a more mixed behavior, e.g., a larger coarse grain diameter should lead to a slower migration and straighter channels, which is not always the case (figure C.3). And some quick tests have shown that those changes vary in intensity depending on the location in the latent space. In addition, we only use averages for the aggradation rate and mean storm rainfall instead of the full time series, leading to a loss of valuable information. From a wider perspective, such label conditioning would allow geologists to more naturally and directly interact with the modeling process. This would help create a feedback loop between geological interpretation and modeling to maximize insight gathering and improve predictions.

Local restructuring of the latent space through pivotal tuning appears to be an effective and robust approach for fine-tuning the match to the data, but only as long as the prior samples are not too far from those data. This remains its key drawback: our tests confirm Roich et al. (2022)'s observation that using random samples instead of pre-inverted samples leads to poor results, so it can only come after an initial, partially successful inversion step. It also means that pivotal tuning can do little to compensate a lack of sample diversity. As mentioned in the previous paragraph, label conditioning is a way forward to get more appropriate pre-inverted samples, although its combination with pivotal tuning still needs to be tested. Latent overparameterization is another way forward, although some quick tests showed that pivotal tuning performs worse with larger latent sizes. We also reached some memory issues, suggesting that wider architecture changes might be necessary to fully explore the benefits of this strategy. From this perspective, some missing features from BigGAN could still improve the inversion, in particular the self-attention layer. But an even bigger change is worth considering: most GAN inversion developments focus on a different architecture called StyleGAN (Karras et al., 2019). StyleGAN's specificity comes from its two latent spaces. The first one, the $\mathcal{Z}$ space, is an input to the network's layers, akin to BigGAN's skip connections for the latent vector used in our bigger GAN (Brock et al., 2019). This input goes through some fully-connected layers, creating a second latent space, the $\mathcal{W}$ space, which is then fed to the convolutional blocks to build an image. The $\mathcal{W}$ space is less entangled, so more amenable to inversion (Karras et al., 2019). The principle is similar to the inference network, except that this is done during training, without any inversion data. Expending StyleGAN or one of its variants (e.g., Karras et al., 2020, 2021; Sauer et al., 2022) to 3D and perform an ablation study to explore its impact on inversion would be valuable. Although the $\mathcal{W}$ space alone is insufficient to get proper inversion results (e.g., Abdal et al., 2019; Zhu et al., 2021; Roich et al., 2022), it might behave like GAN training (Rongier & Peeters, 2025c): datasets used in deep learning are – at least for now – more complex than their counterparts in geology, so that advanced techniques developed by the deep-learning community might work directly.

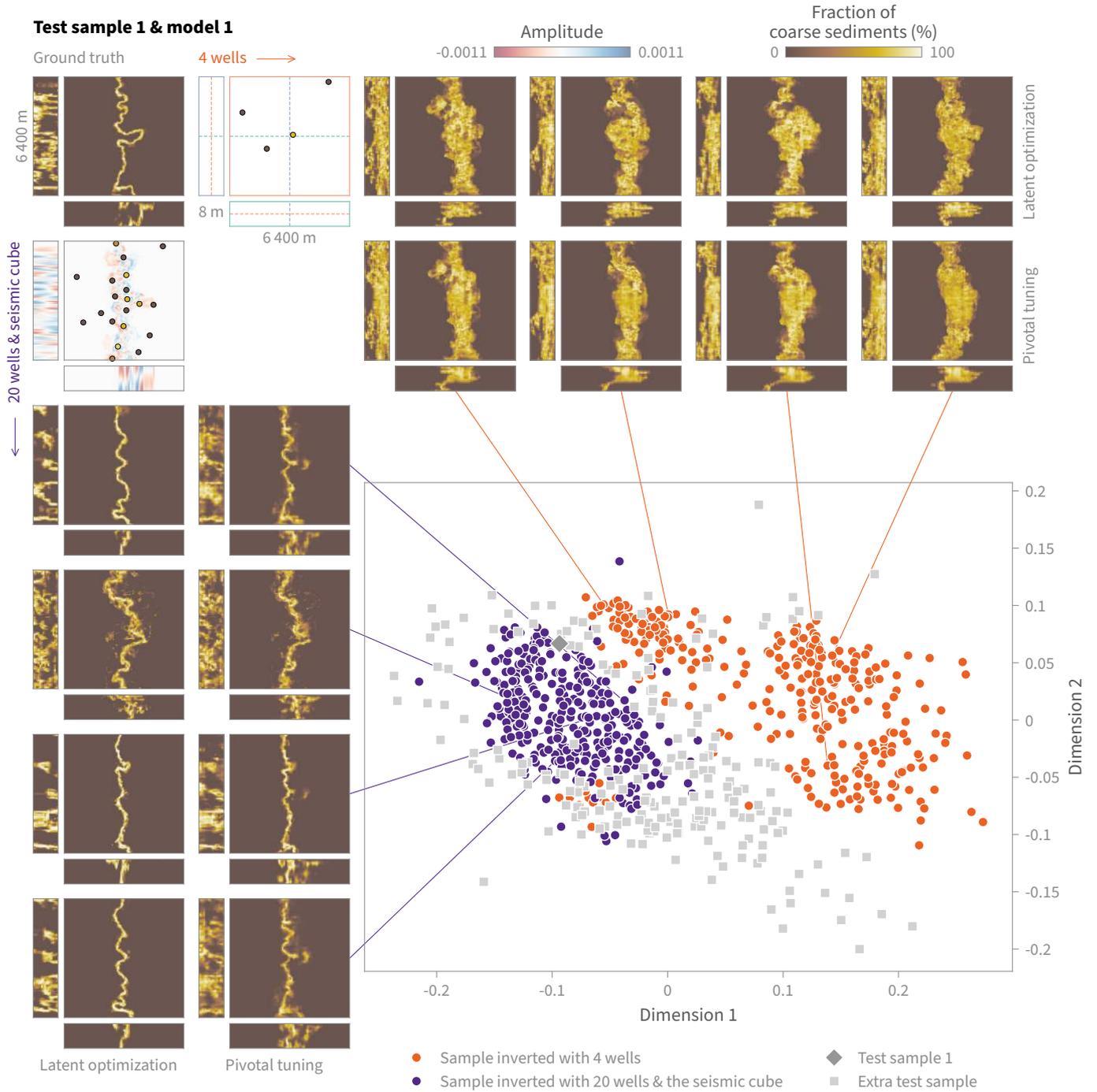

**Figure 12** Testing sample quality and diversity from model 1 using mutlidimensional scaling to represent the sliced Wasserstein distances between 200 test samples and 300 samples inverted using pivotal tuning based on 4 wells or 20 wells and the seismic cube from test sample 1. The samples shown were randomly selected. The slices go through the center of the samples, as shown on the data with 4 wells.

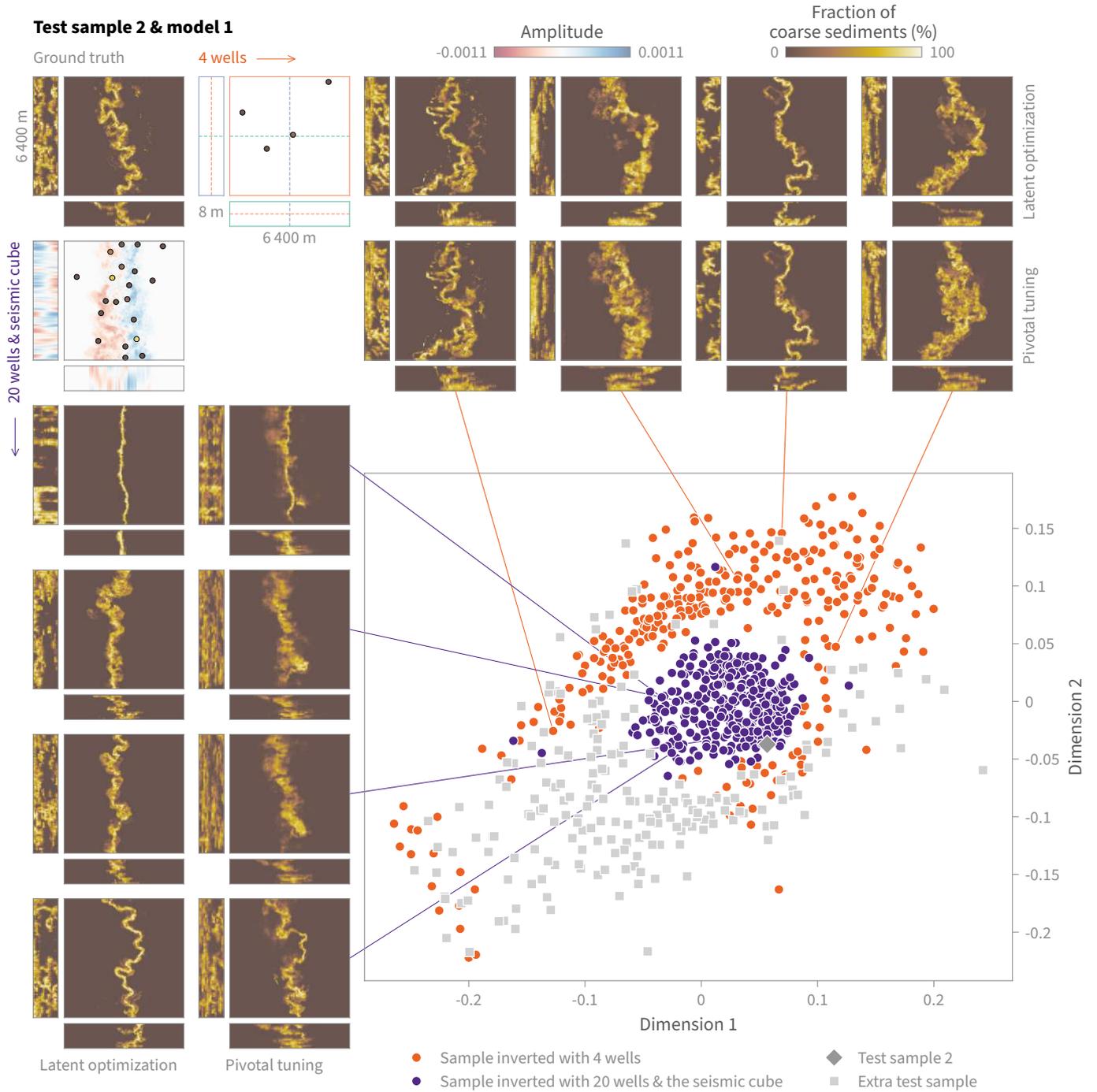

**Figure 13** Testing sample quality and diversity from model 1 using mutlidimensional scaling to represent the sliced Wasserstein distances between 200 test samples and 300 samples inverted using pivotal tuning based on 4 wells or 20 wells and the seismic cube from test sample 2. The samples shown were randomly selected. The slices go through the center of the samples, as shown on the data with 4 wells.

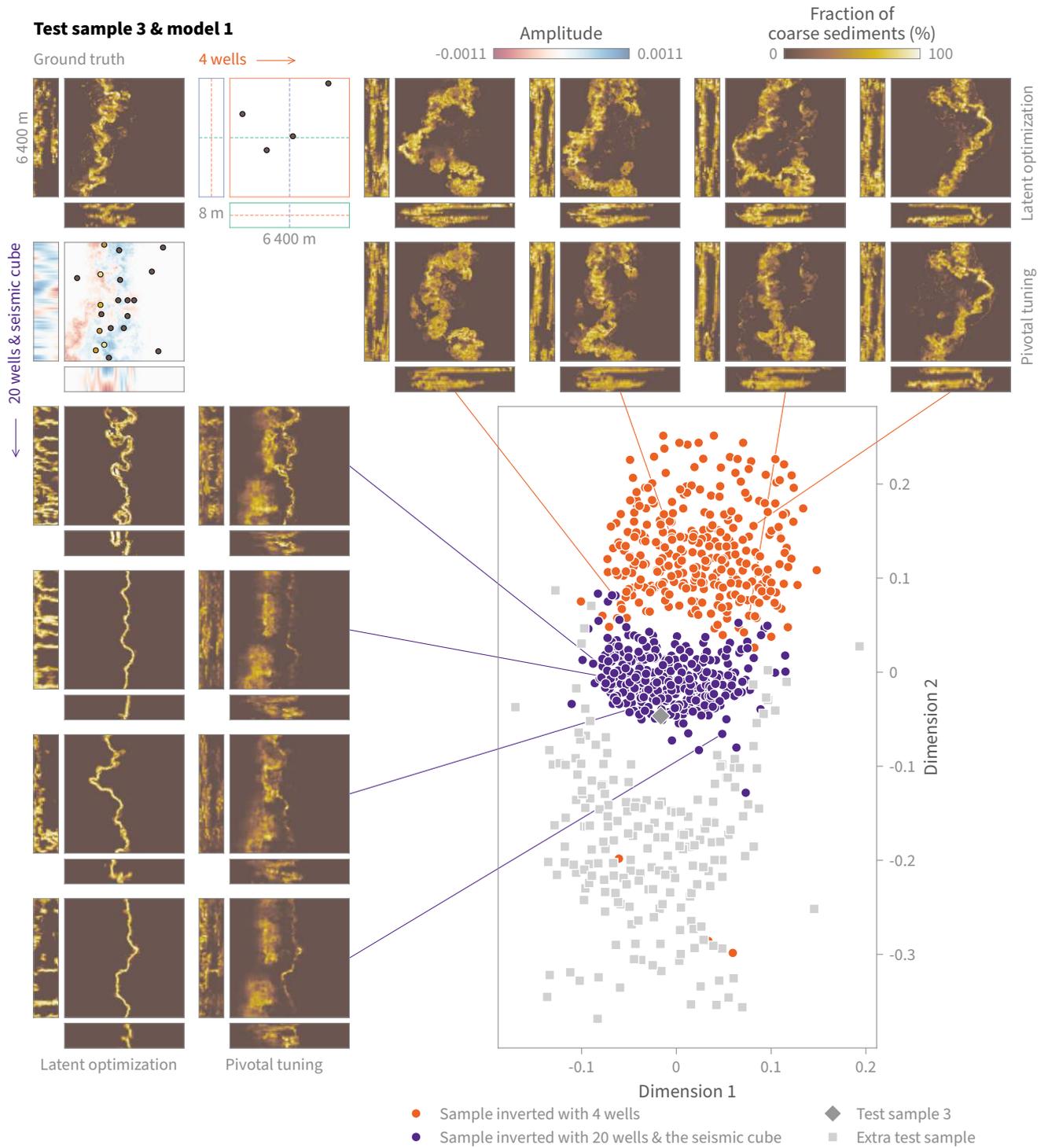

**Figure 14** Testing sample quality and diversity from model 1 using mutlidimensional scaling to represent the sliced Wasserstein distances between 200 test samples and 300 samples inverted using pivotal tuning based on 4 wells or 20 wells and the seismic cube from test sample 3. The samples shown were randomly selected. The slices go through the center of the samples, as shown on the data with 4 wells.

If the opposite holds, alternative approaches might prove more practical. GAN conditioning is the most obvious one. It has already been successfully tested on stationary discrete properties in 2D and in 3D with two facies, and multiple wells and seismic-derived data (Song, Mukerji, & Hou, 2022; Song, Mukerji, Hou, et al., 2022). Testing it on the case studies from this work would lead to a more direct and comprehensive comparison of this strategy's performance. For larger models trained on more diverse datasets, training could also be done in two steps for efficiency: a first unconditional step for generic pretraining, and a second conditional step for matching specific data types. This would be similar to how multimodal large language models are trained (e.g., Dai et al., 2024; Deitke et al., 2025; Grattafiori et al., 2024). This would also be similar to pivotal tuning's principle, except that pivotal tuning does not need any of the training data, only the pretrained GAN. Another alternative is to turn to an invertible deep generative model, for instance a variational autoencoder (VAEs) (Kingma & Welling, 2014; Higgins et al., 2017) or a flow-based generative model (Asim et al., 2020).

Like most studies applying GANs to subsurface modeling, our case studies are entirely synthetic. Even if we made them as plausible as possible, and even if having a ground truth comes with advantages, testing GANs on real data should be the next target. Having test sample 3 poorly represented in the training data already made inversion difficult; turning to a real case will make matters worse. Approaches like the inference network or pivotal tuning can be seen as a posteriori curation of the training space, but they cannot expand beyond that training space. We then need a clearer perspective on to properly and efficiently design training datasets from process-based models for such tasks. Turning to real data will also imply uncertainties in the data-generating process, as well as potential biases. This is especially true for seismic data. It is then unclear whether a technique like pivotal tuning can still reach the 1 % threshold in those conditions.

# 5 Conclusions

In geological modeling as well as in the more general field of deep learning, inversion has been the most explored approach to match the samples of a GAN to some data. Most of those approaches rely on sampling from the GAN's latent space with some process to reduce the mismatch between samples and data. While they have been shown to work on simple 2D cases related to geological modeling, they tend to struggle as complexity increases, and end up with a big data mismatch or seldom any sample diversity. But the key bottleneck lies more with the internal representation that GANs learn from the data – and express through their latent space – than with the inversion approaches themselves. Indeed, GANs are prone to construct entangled latent spaces, which likely lack geological consistency and make inversion difficult. Disentangling the latent space during the training of the GAN itself is then key to improve inversion performances. Approaches to restructure the latent space after a partially successful first inversion step – like pivotal tuning – can still reduce data mismatch to acceptable levels. Since GANs encode stronger geological priors, they can go around biases in data sampling to capture the non-stationarity of fluvial deposits. They can also invert geophysical data – removing the need for an intermediate transformation of a seismic amplitude to a sand probability cube for instance – although this aspect needs further testing on more complex cases beyond a single channel belt and a perfectly known data-generating process. While we can extract some rough insights on deposition time without any direct data, inconsistencies and uncertainties need to be reduced for the exercise to have practical value.

Our results show that GANs could already be integrated in the current geological modeling workflow: existing approaches for process-based modeling, GAN training, and GAN inversion are capable of handling all the necessary steps. We now need to move towards more detailed studies on larger case studies closer to field applications to further assess how the whole workflow performs. Unlike geostatistical approaches, which always require some feature engineering, GANs and other deep generative models can shift much of this work to the initial development phase. This requires to train models valid for as many geological settings and potential applications as possible. We can only get there with substantial investments in large and open datasets of geologically plausible synthetic subsurface analogs. This, in turn, will require significant developments in process-based modeling, because developing models for a large variety of sedimentary environments is only one component of a larger system: we need them to be integrated with other models for all the processes shaping the subsurface, such as deformation, fracturing and faulting, and diagenesis.

**Data and software availability**

Our entire study is openly available, including the scripts to reproduce results and figures, the results themselves, the pretrained models (Rongier & Peeters, 2025b), and the training data (Rongier & Peeters, 2021). Those scripts uses the open-source Python package voxgan (Rongier, 2021), which is built upon PyTorch (Ansel et al., 2024). The figures were made using the open-source Python packages matplotlib (Hunter, 2007) and PyVista (Sullivan & Kaszynski, 2019).

## Acknowledgments

We would like to thank the GSE Computational Team for their support, and Frie Van Bauwel for her valuable feedback.

# Appendix A  Detailed data-generating process

Well locations were randomly selected based on a two-step process:

1. In a first step, 4 vertical wells were selected sequentially so that each well had a 1 km exclusion zone and a zone between 1 and 2 km where the selection probability increased linearly. The selection was also weighted by the fraction of coarse sediments of all three test samples. This ensured that the wells are spread out over the samples with at least one well going through some coarse sediments. Those 4 wells are then common for all three test samples, and represent some legacy wells that were targeting different intervals.

2. In a second step, 16 more vertical wells were selected sequentially so that each well had a 0.5 km exclusion zone and a zone between 0.5 and 1 km where the selection probability increased linearly. In this step, the legacy wells had a 0.25 km exclusion zone and a zone between 0.25 and 0.5 km where the selection probability increased linearly. The selection was also weighted by the fraction of coarse sediments but for each test sample separately. Those new wells are then different for each test sample, and represent wells that target these specific coarse deposits.

Each well only contains the fraction of coarse sediments, not the deposition time, directly extracted from the test samples. We then defined three well datasets for each test sample: one with the 4 legacy wells, one with 8 wells, and one with 20 wells (figure 1).

Seismic cubes were generated based on a mixture of coarse and fine sediments that remain unconsolidated after burial. The petrophysical model assumes that the coarse fraction is made of quartz with a density of $2.65\,\mathrm{g\,cm^{-3}}$, a porosity of $27\%$, a bulk modulus of $37\,\mathrm{GPa}$, and a shear modulus of $44\,\mathrm{GPa}$; the fine fraction is made of clay with a density of $2.6\,\mathrm{g\,cm^{-3}}$, a porosity of $14\%$, a bulk modulus of $21\,\mathrm{GPa}$, and a shear modulus of $7\,\mathrm{GPa}$; and the fluid is made of water with a density of $1\,\mathrm{g\,cm^{-3}}$ and a bulk modulus of $2.29\,\mathrm{GPa}$. We used the constant-clay model for shaly sands to compute the dry bulk and shear moduli and the density for the mixed lithology (assuming a critical porosity of $50\%$ and an effective pressure of $10\,\mathrm{MPa}$), then Gassmann's equation to compute the water-saturated moduli, finally the water-saturated moduli and density to compute the P-wave velocity (Avseth et al., 2005). From the P-wave velocity and the density, we computed impedance and reflectivity at normal incidence. Our samples are thin compared to the vertical resolution of seismic data, so we added a structureless over- and under-burden and our seismic cube has 51 cells along the vertical direction. Finally, we simulated a 3D seismic prestack depth migration image – i.e., a seismic cube (figure 1) – for each of our test samples using a 3D convolution approach based on a point-spread function (Lecomte, 2008; Lecomte & Kaschwich, 2008; Lecomte et al., 2015). We used a Ricker wavelet with a peak frequency of $60\,\mathrm{Hz}$ – slightly higher than conventional values to keep the inversion tests meaningful – an incident angle of $0°$, and an illumination angle of $45°$.

# Appendix B  Progression of the initial GAN inversions

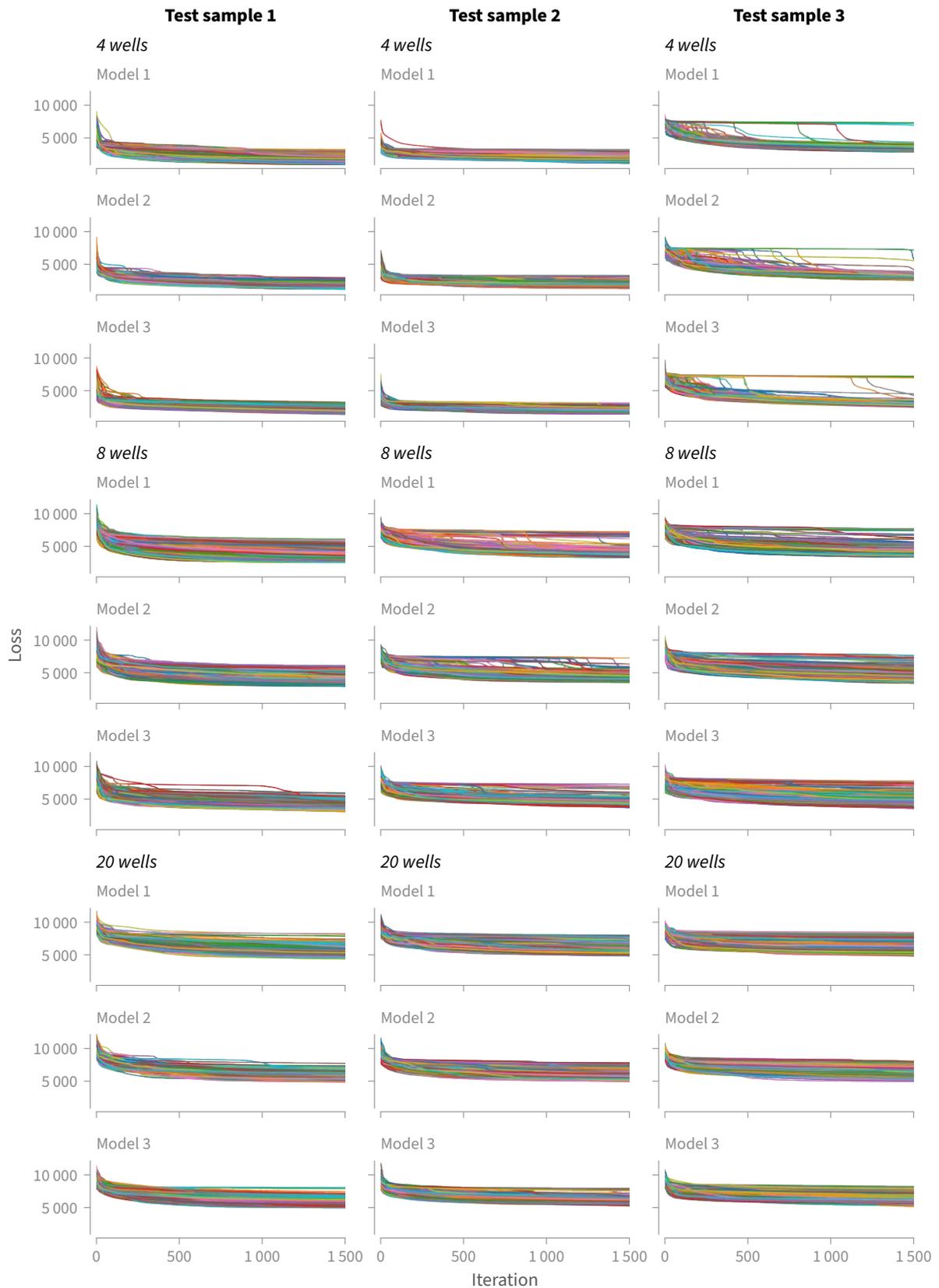

**Figure B.1**  Losses during the latent optimization to invert 300 GAN samples.

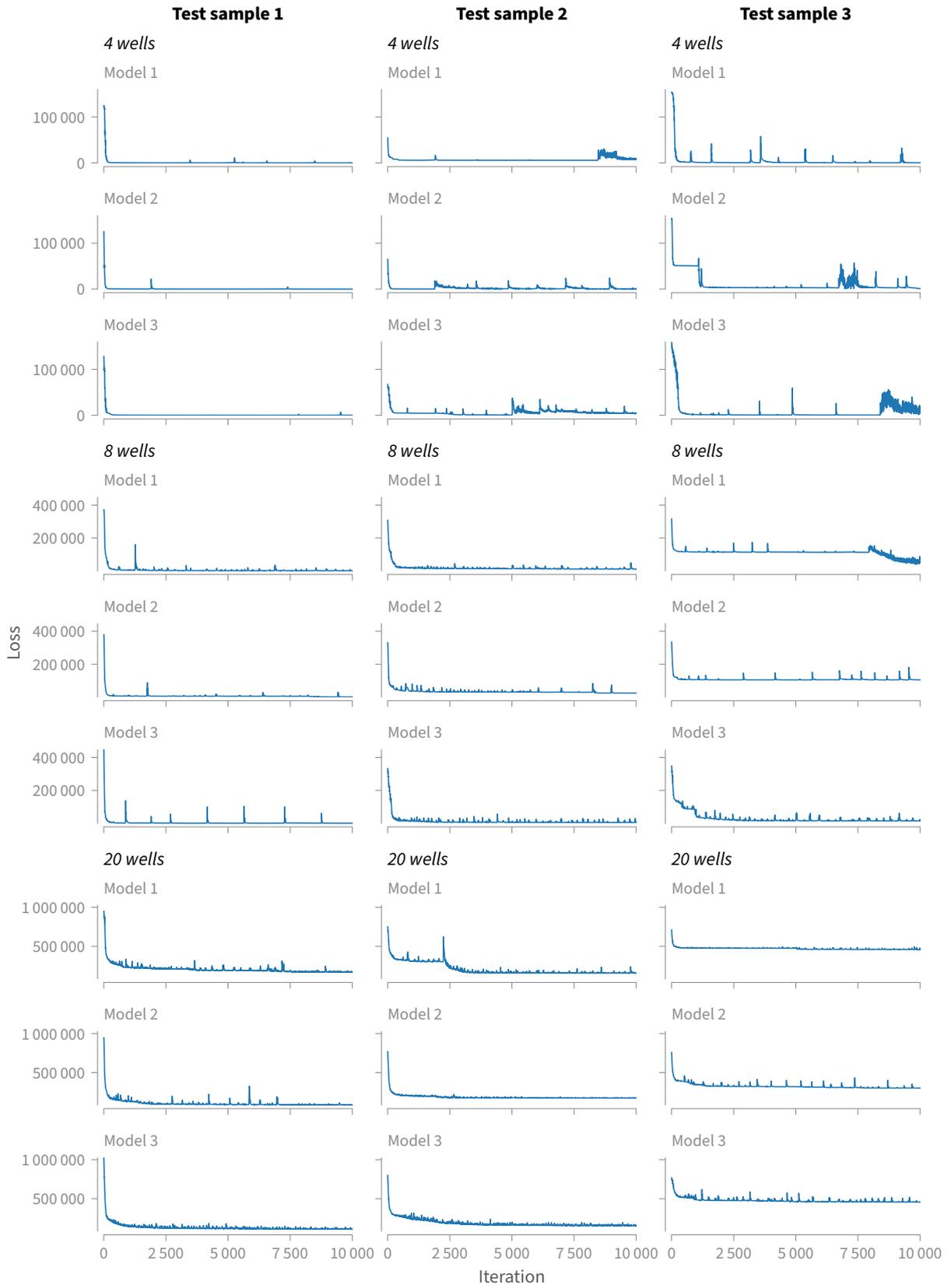

**Figure B.2** Training loss of the inference network.

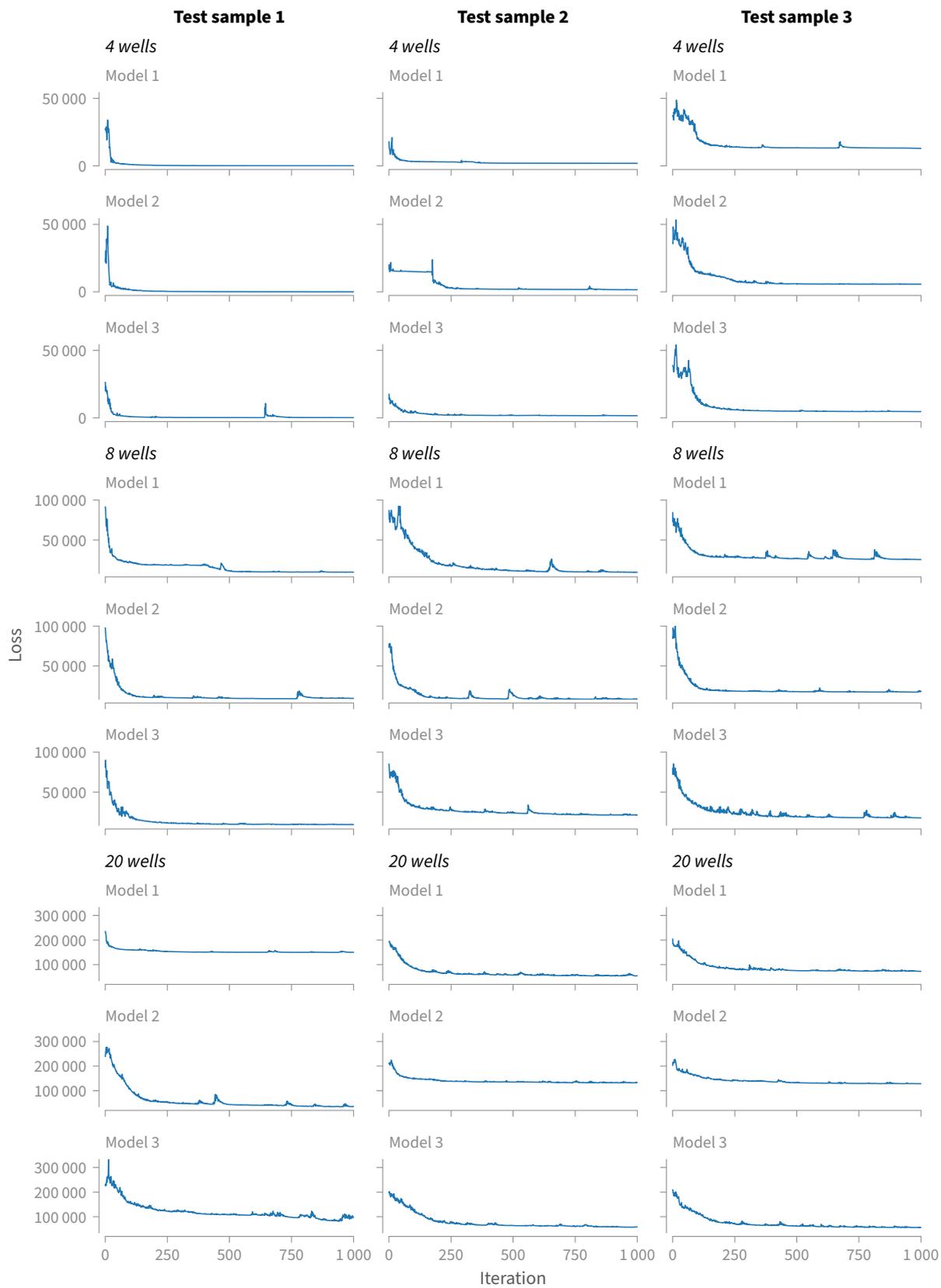

**Figure B.3** Variational inference loss.

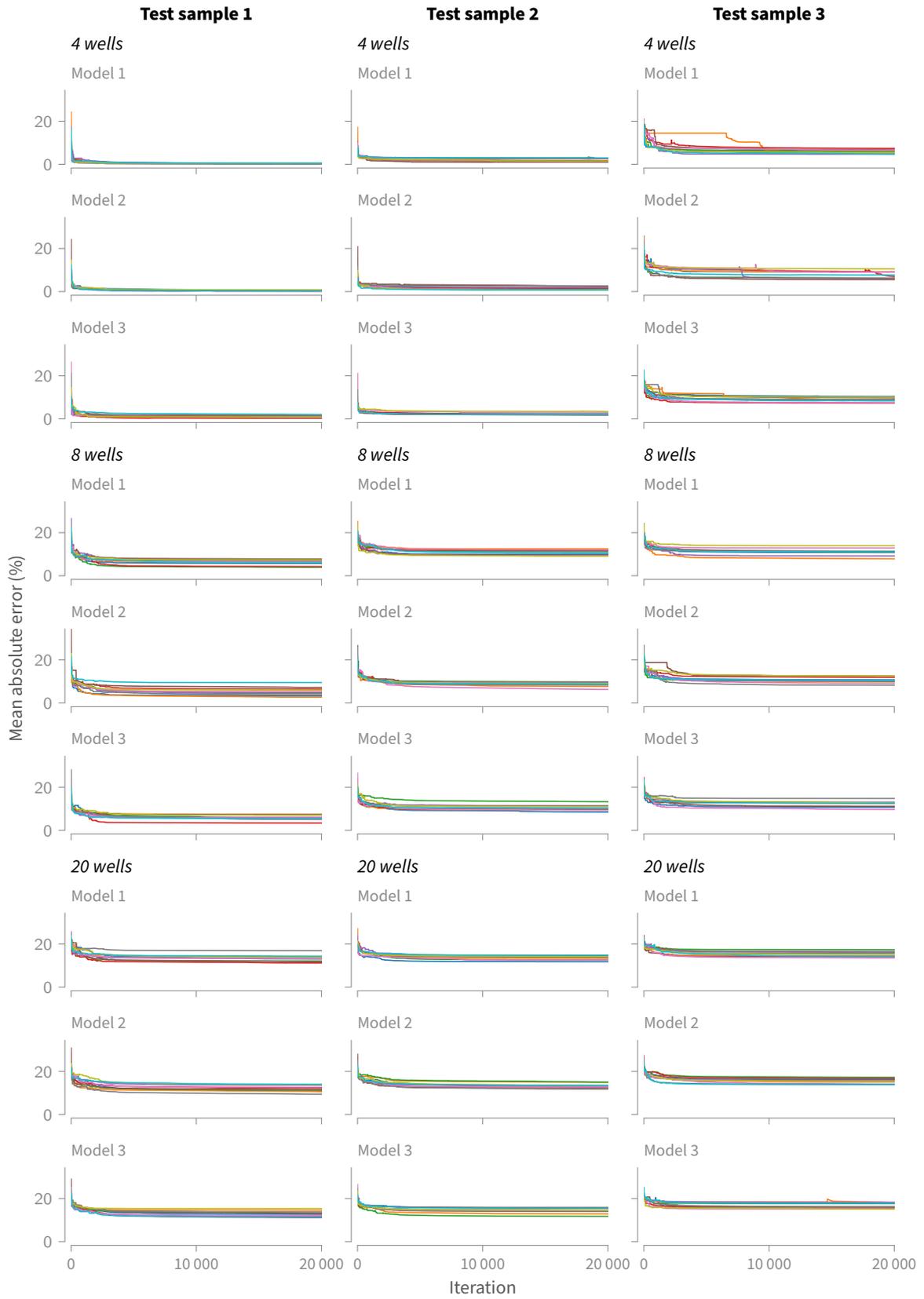

**Figure B.4**   Inversion error for the ten chains during sampling with DREAM$_{(ZS)}$, including the 20 000 steps of burn-in.

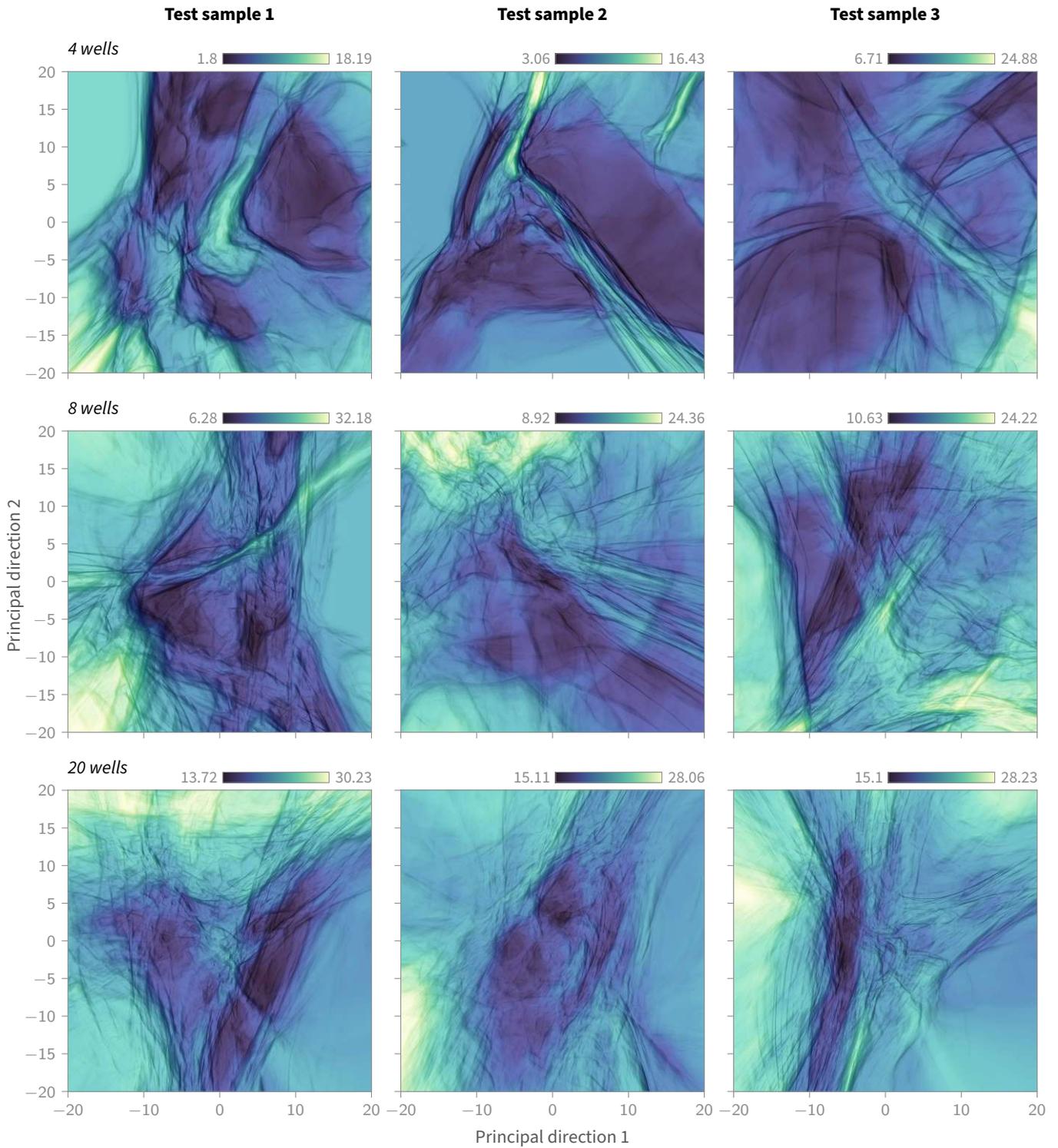

**Figure B.5** Inversion error landscape for model 2.

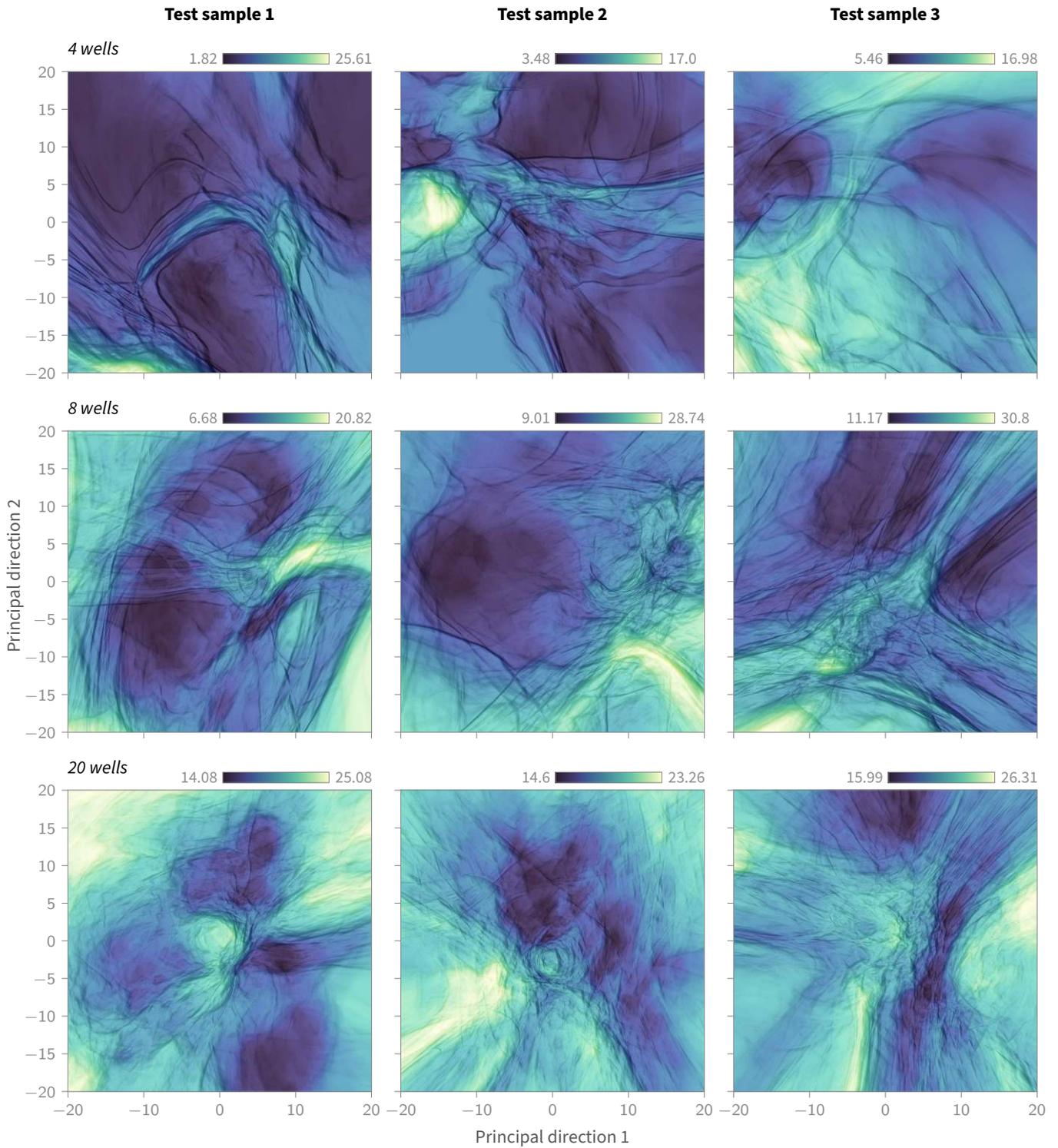

**Figure B.6**  Inversion error landscape for model 3.

# Appendix C  Inversion progression with latent restructuring

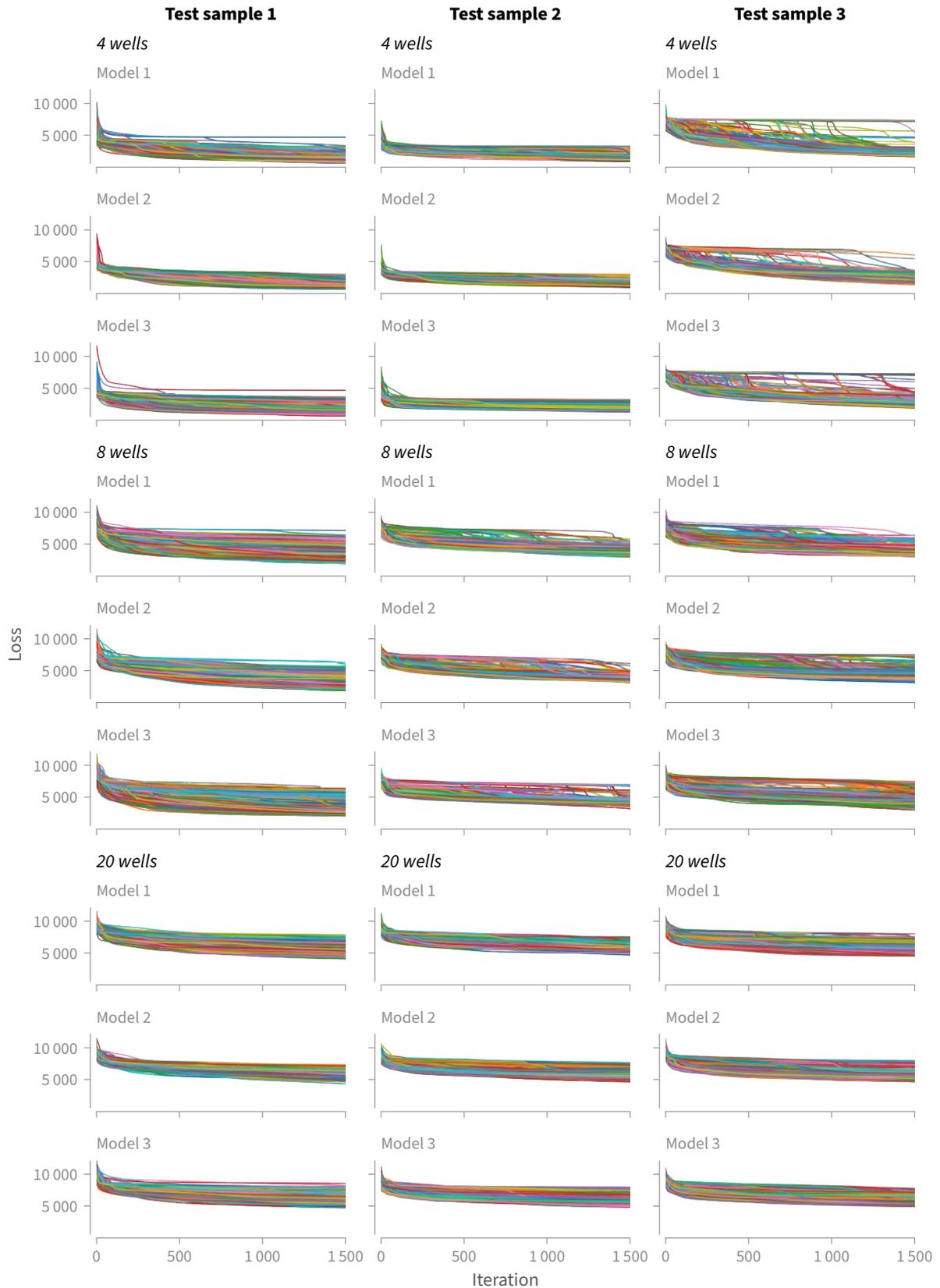

**Figure C.1**  Losses during the latent optimization to invert 300 GAN samples from a model closer to BigGAN.

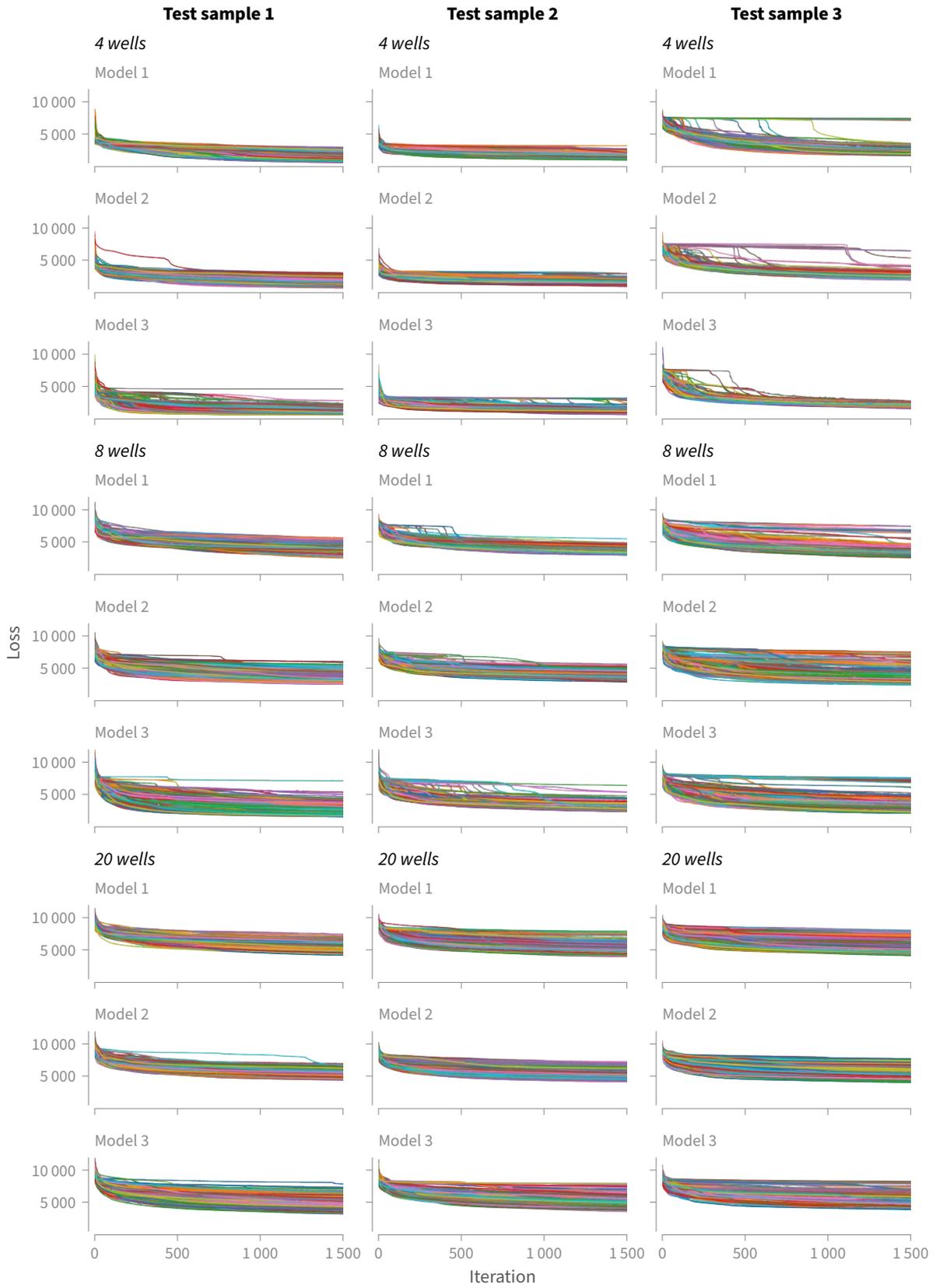

**Figure C.2** Losses during the latent optimization to invert 300 GAN samples from a model with latent size of 512.

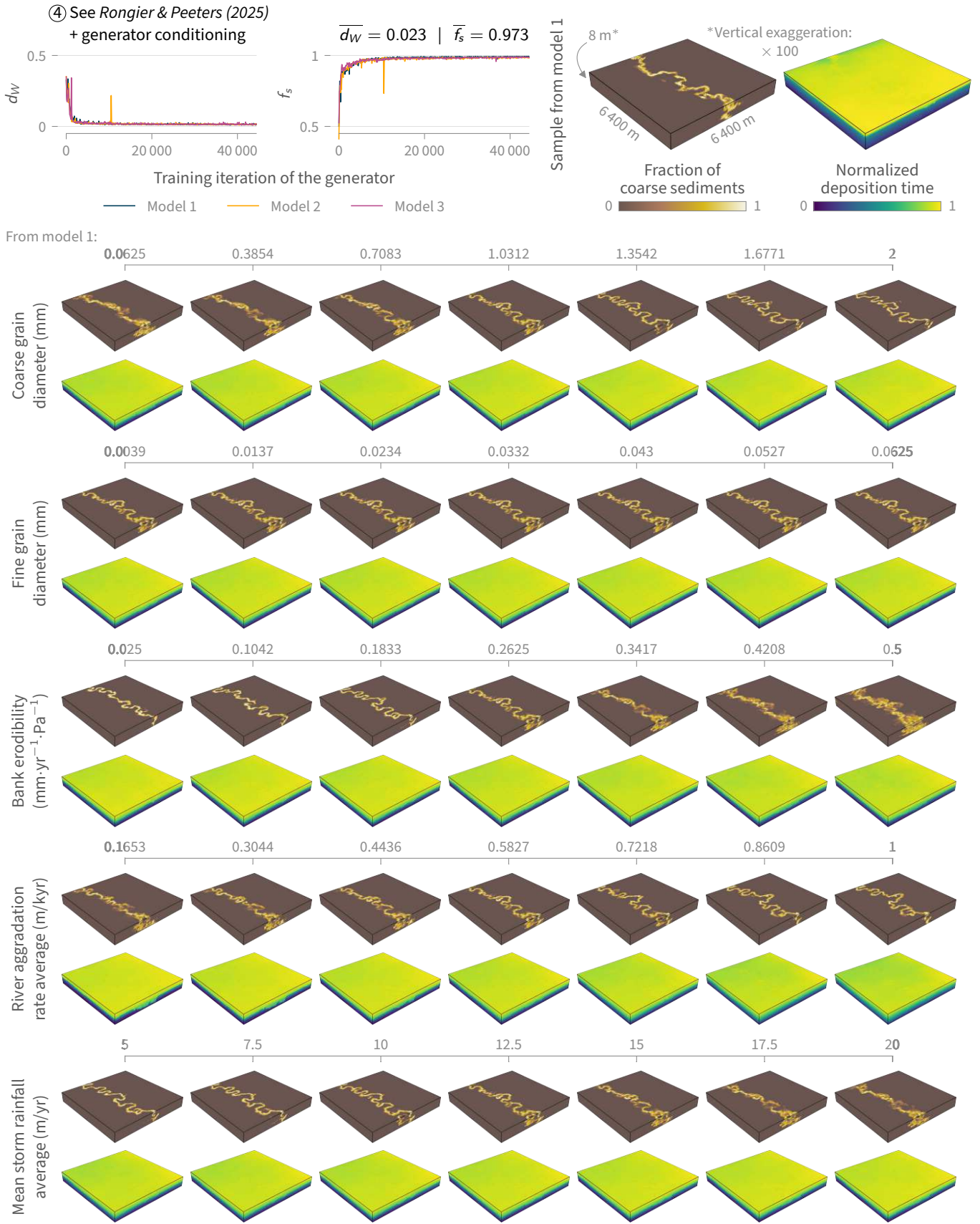

**Figure C.3** Training of a conditional GAN based on architecture 4 from Rongier and Peeters (2025c) and interpolation along the labels with model 1. $d_W$: sliced Wasserstein distance to the validation set; $f_s$: fraction of cells honoring the law of superposition.

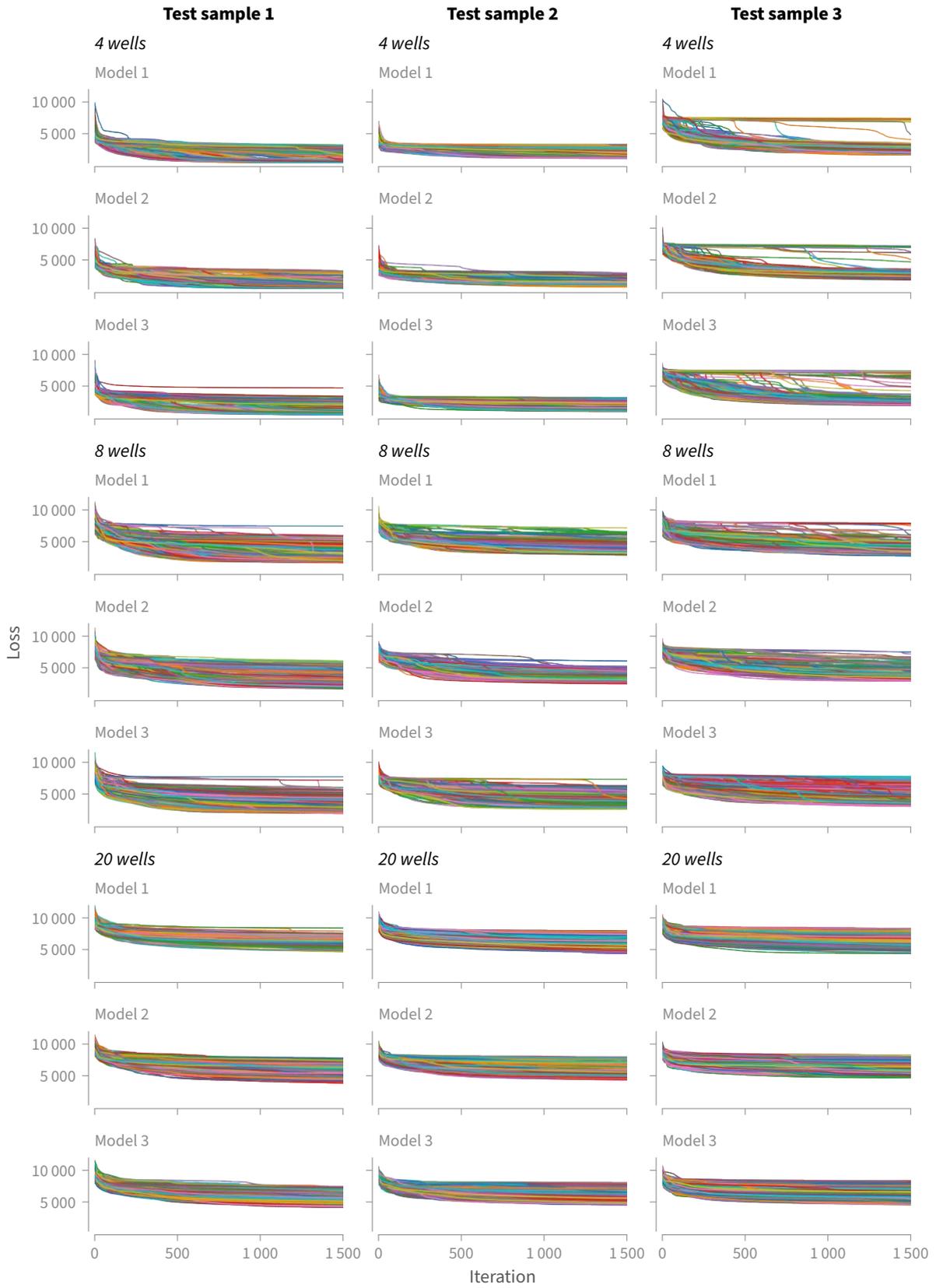

**Figure C.4** Losses during the latent optimization to invert 300 GAN samples from a model with conditioning.

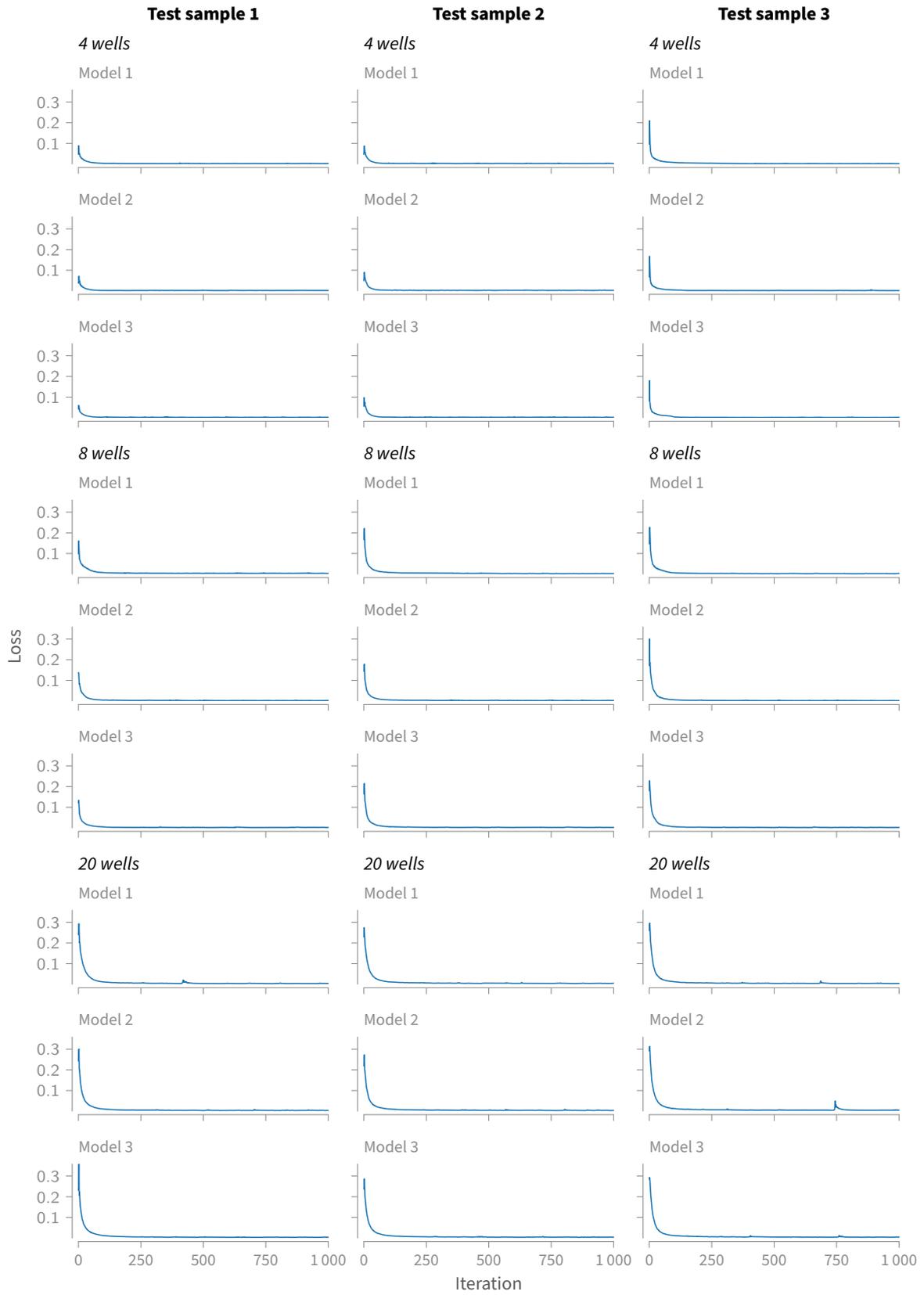

**Figure C.5** Training loss of the pivotal tuning.

# Appendix D  Inversion progression with seismic data

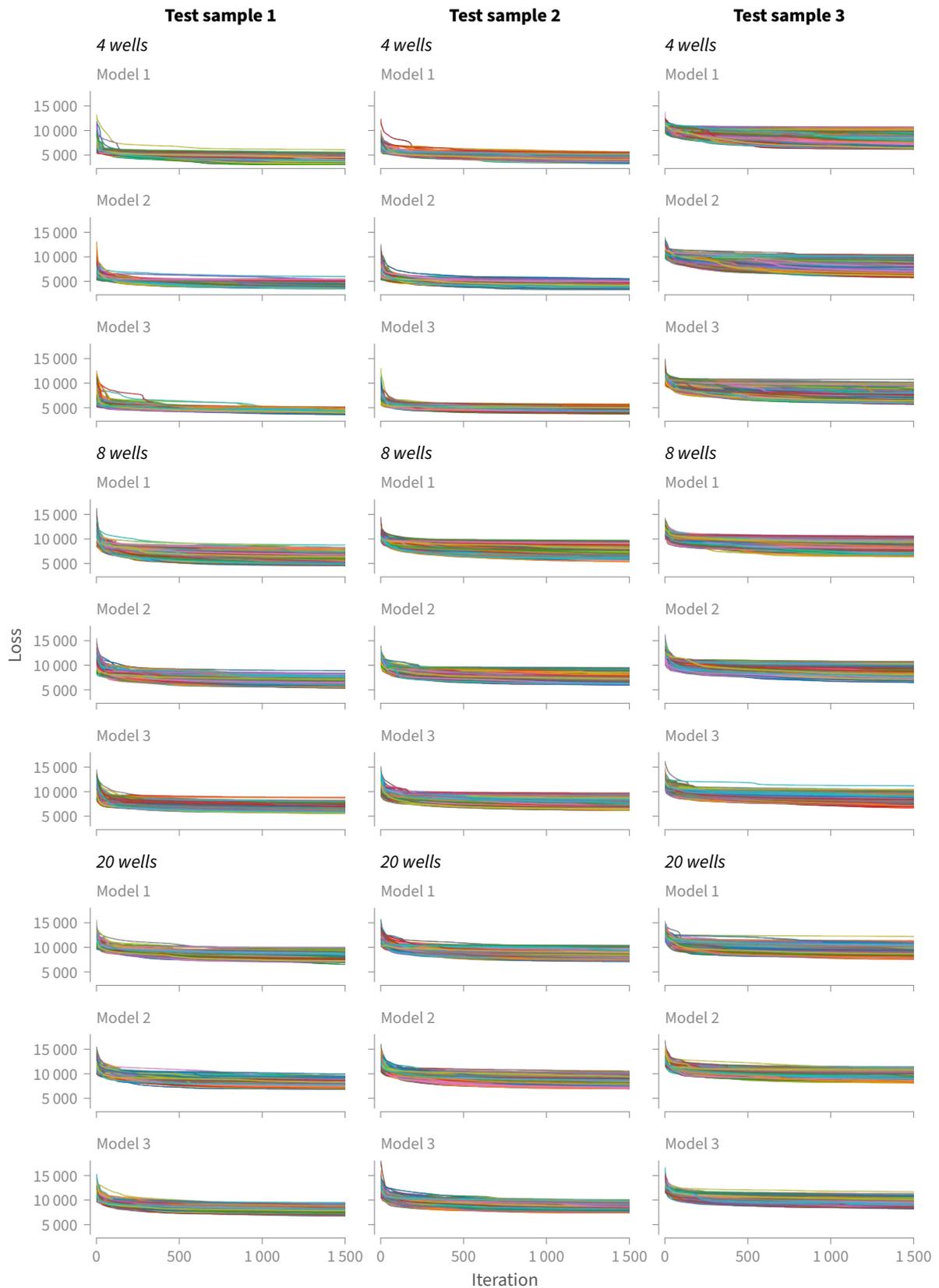

**Figure D.1**  Losses during the latent optimization to invert 300 GAN samples with seismic data.

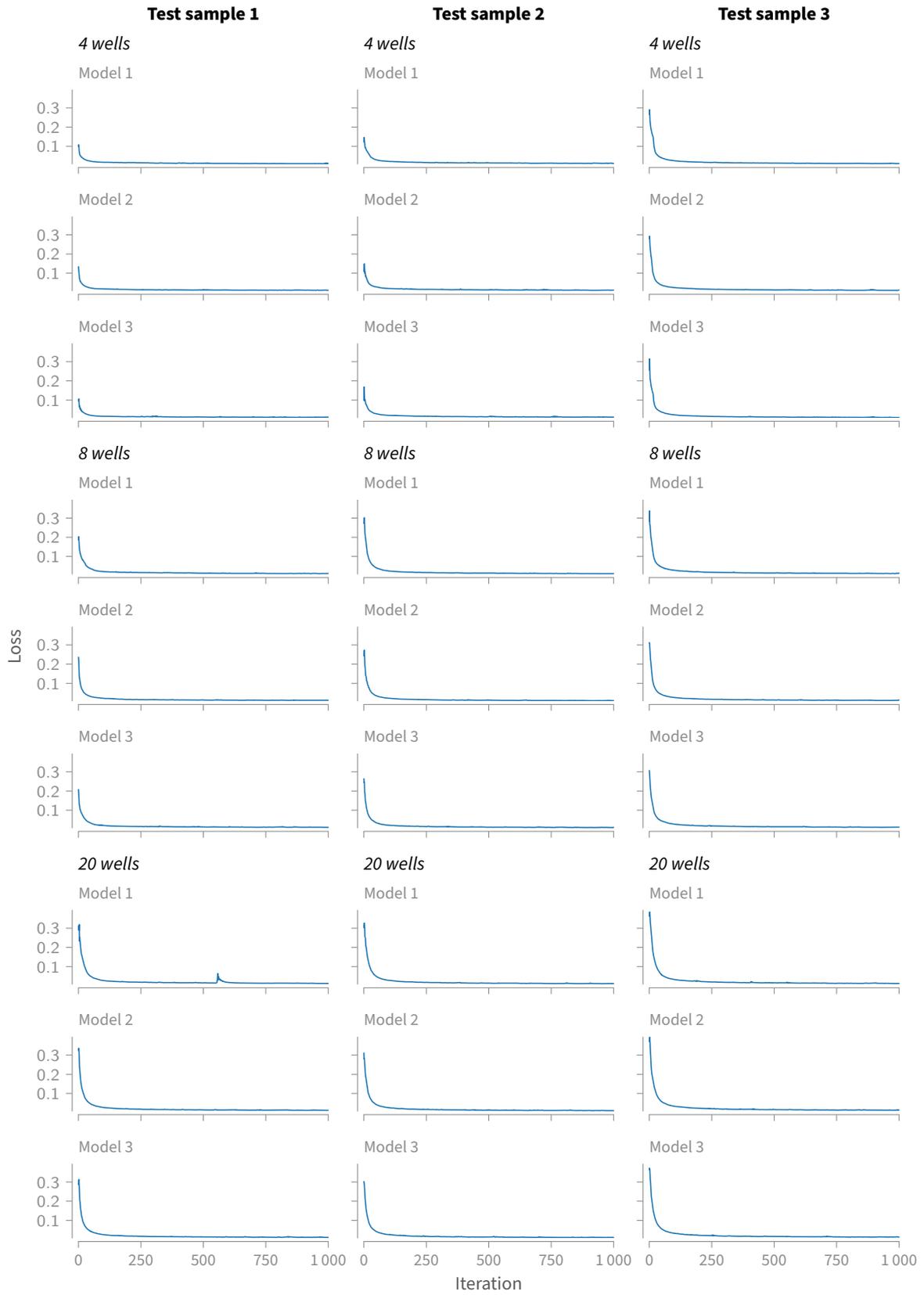

**Figure D.2** Training loss of the pivotal tuning with seismic data.

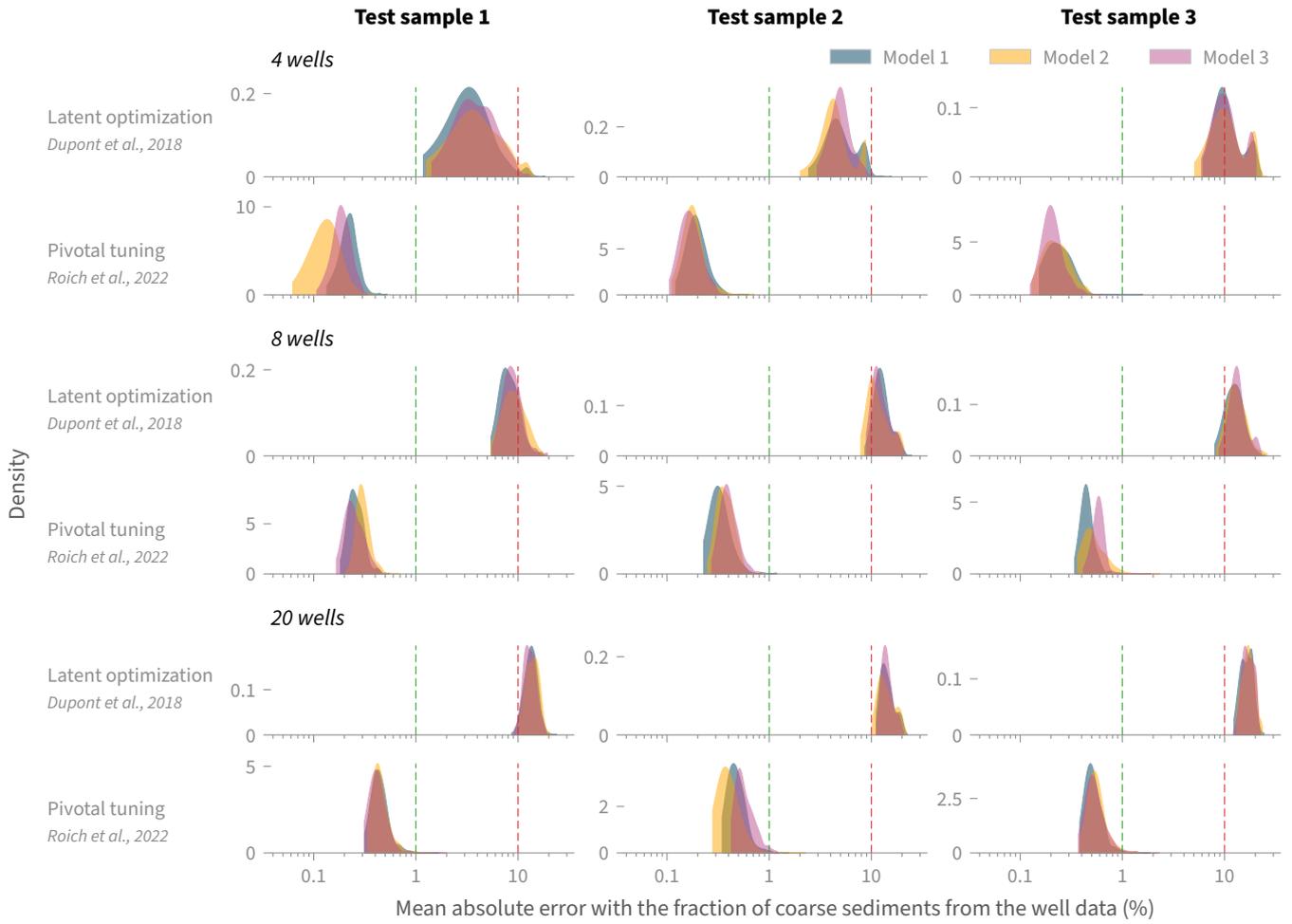

**Figure D.3** Inversion error between 300 inverted GAN samples and the well data for the latent optimization and the pivotal tuning with seismic data.

# Appendix E  Samples after inversion

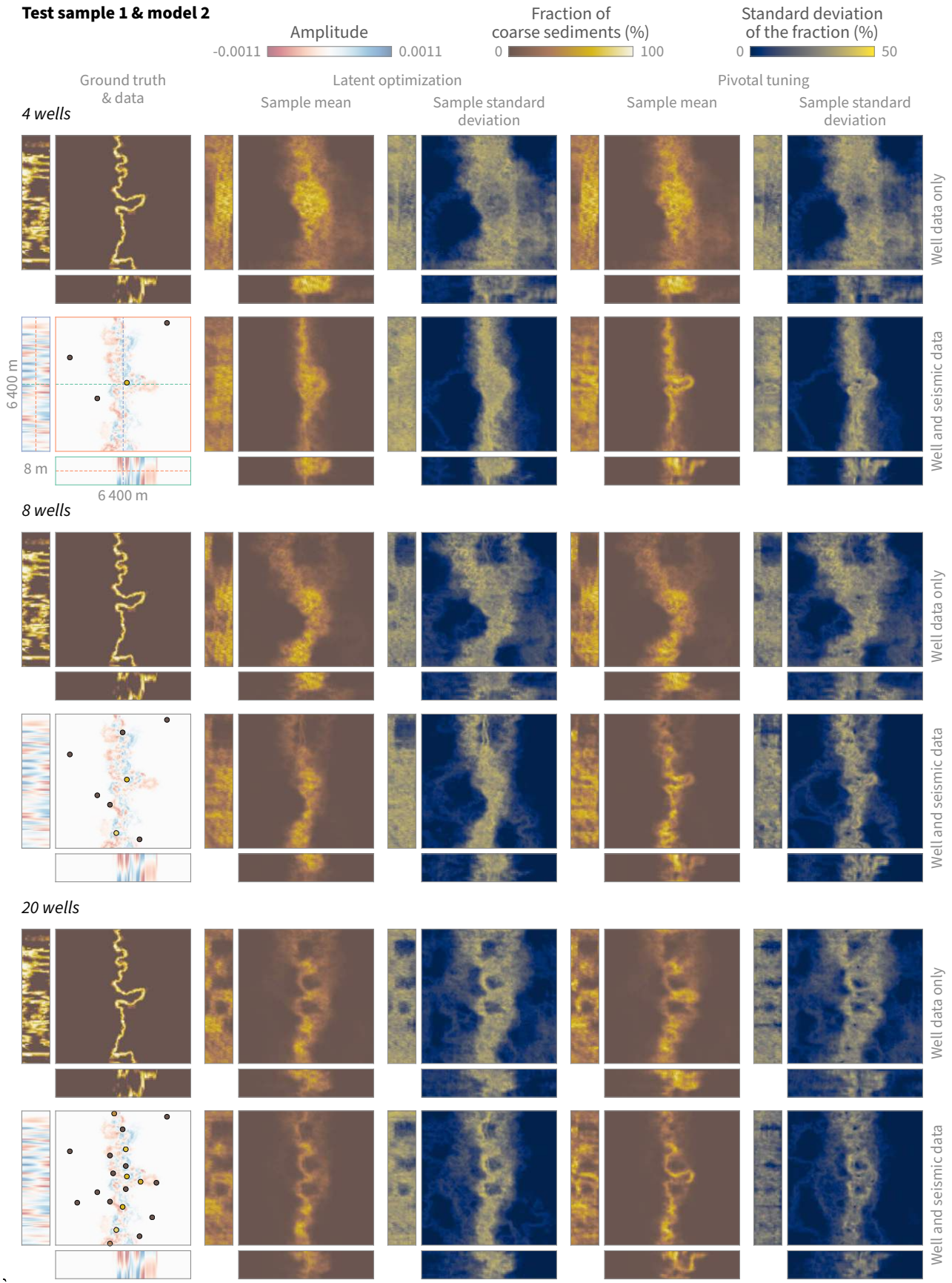

**Figure E.1** Comparison between the mean and standard deviation of the fraction of coarse sediments for 300 samples inverted from the GAN model 2 with test sample 1 for two inversion approaches without and with seismic data. The slices go through the center of the samples, as shown on the data with 4 wells.

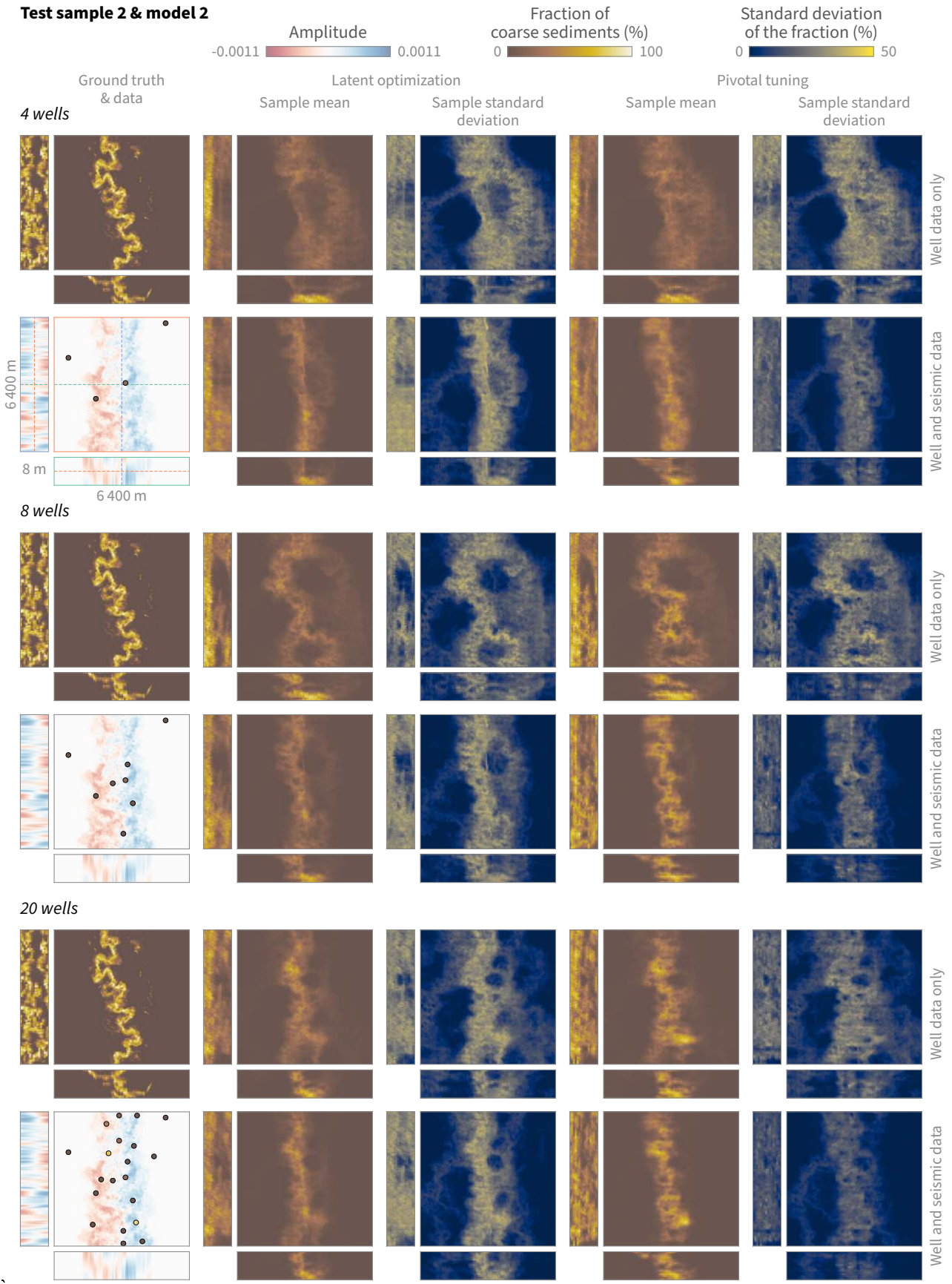

**Figure E.2** Comparison between the mean and standard deviation of the fraction of coarse sediments for 300 samples inverted from the GAN model 2 with test sample 2 for two inversion approaches without and with seismic data. The slices go through the center of the samples, as shown on the data with 4 wells.

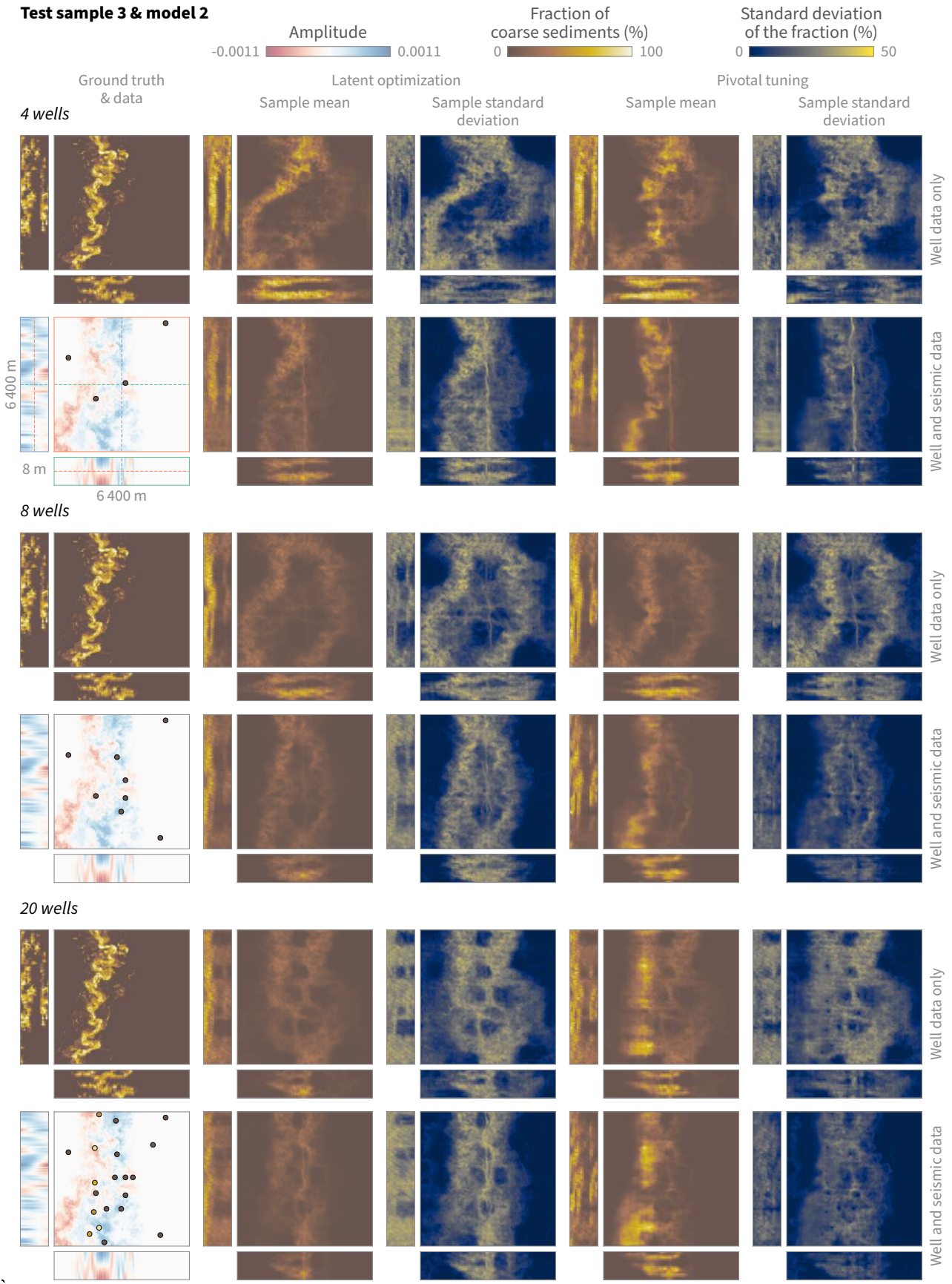

**Figure E.3** Comparison between the mean and standard deviation of the fraction of coarse sediments for 300 samples inverted from the GAN model 2 with test sample 3 for two inversion approaches without and with seismic data. The slices go through the center of the samples, as shown on the data with 4 wells.

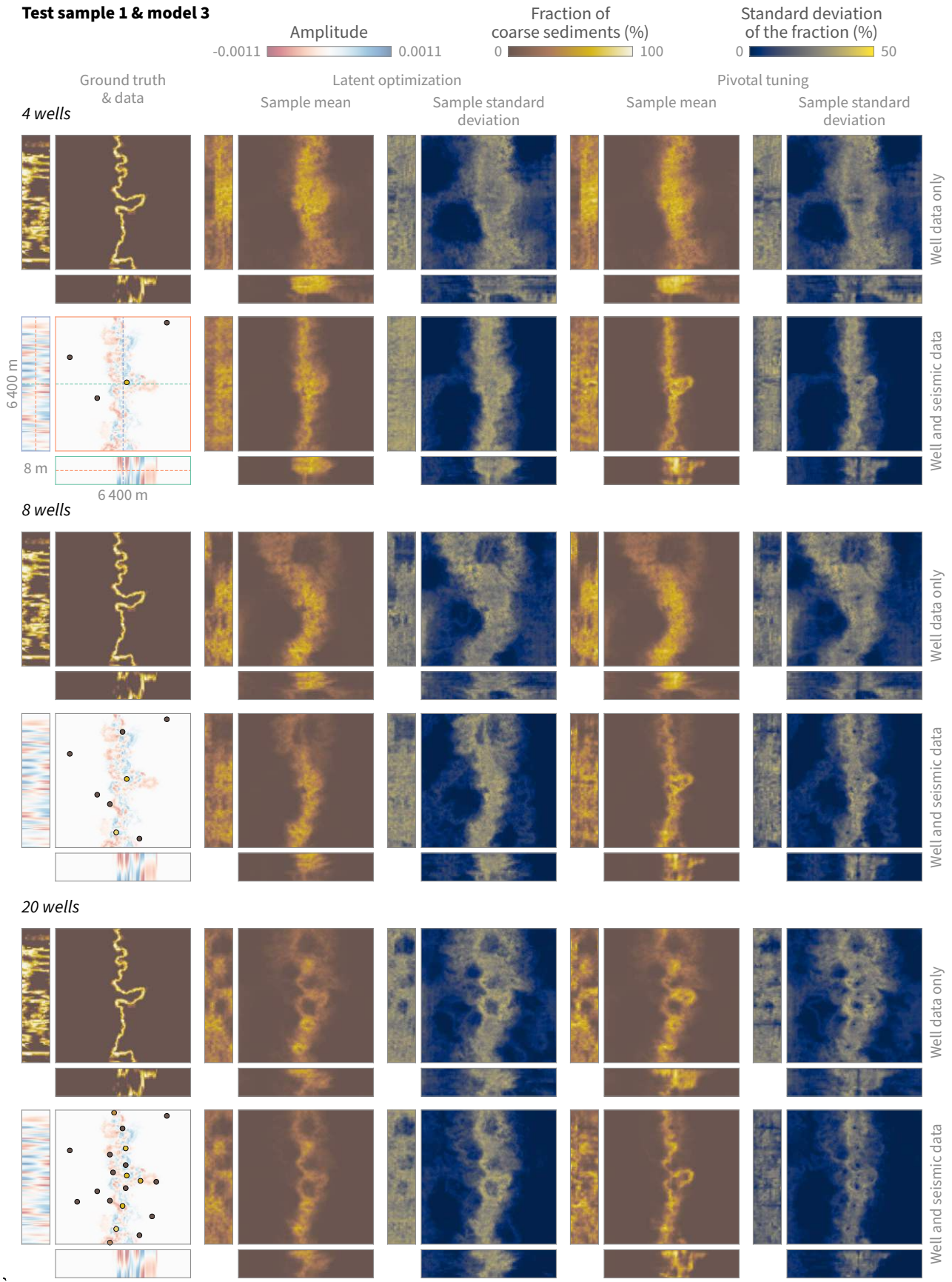

**Figure E.4** Comparison between the mean and standard deviation of the fraction of coarse sediments for 300 samples inverted from the GAN model 3 with test sample 1 for two inversion approaches without and with seismic data. The slices go through the center of the samples, as shown on the data with 4 wells.

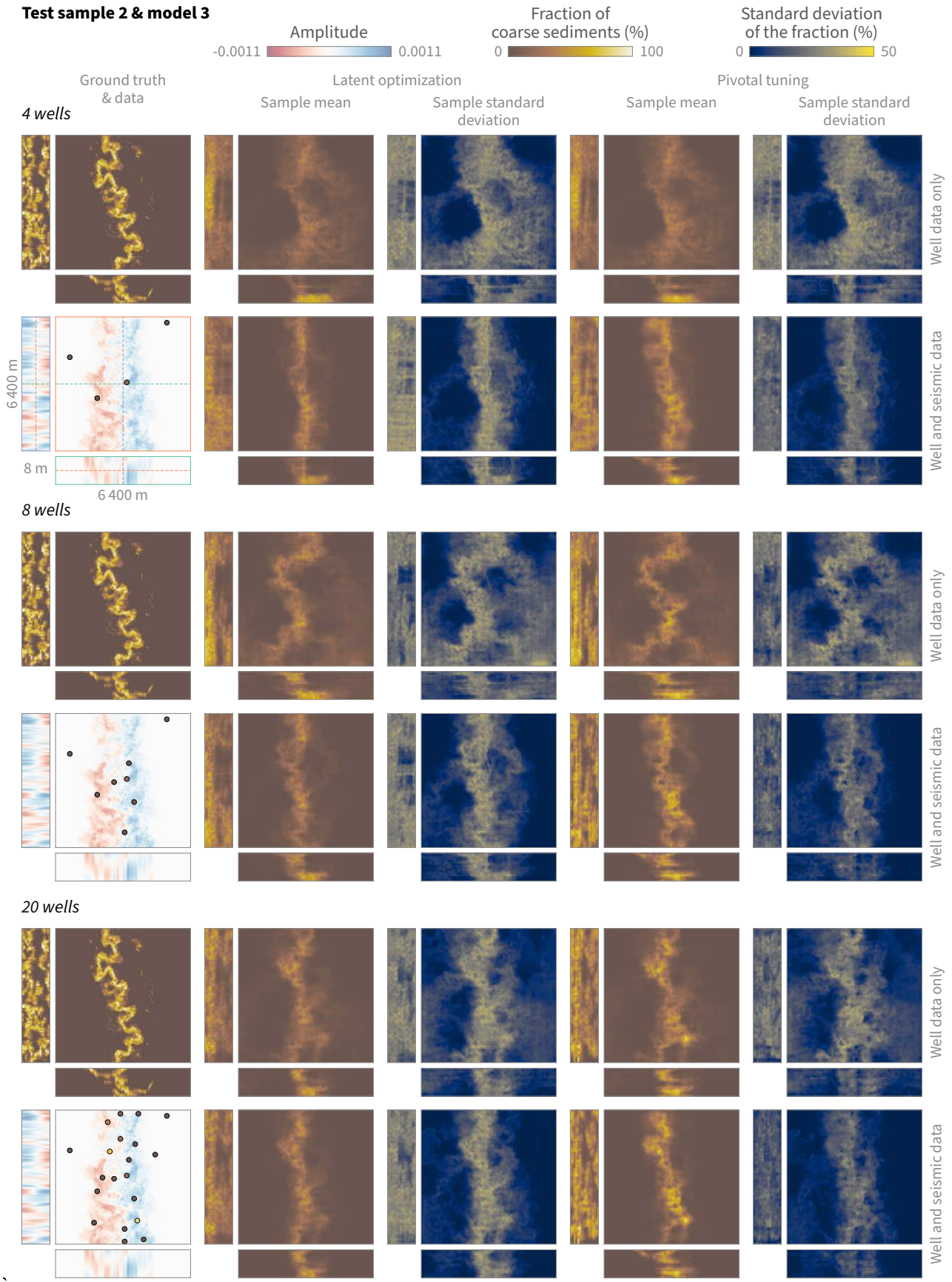

**Figure E.5** Comparison between the mean and standard deviation of the fraction of coarse sediments for 300 samples inverted from the GAN model 3 with test sample 2 for two inversion approaches without and with seismic data. The slices go through the center of the samples, as shown on the data with 4 wells.

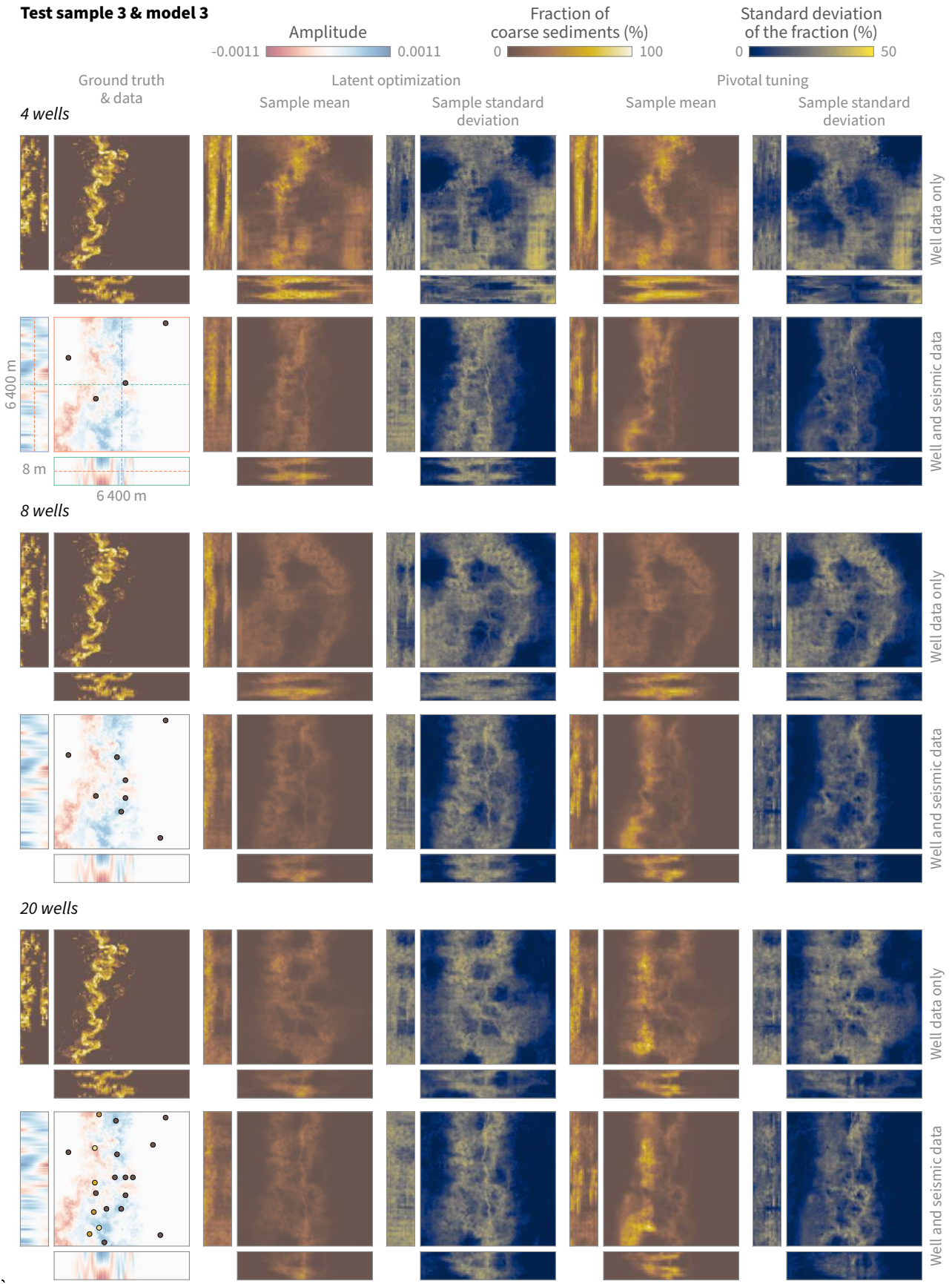

**Figure E.6** Comparison between the mean and standard deviation of the fraction of coarse sediments for 300 samples inverted from the GAN model 3 with test sample 3 for two inversion approaches without and with seismic data. The slices go through the center of the samples, as shown on the data with 4 wells.

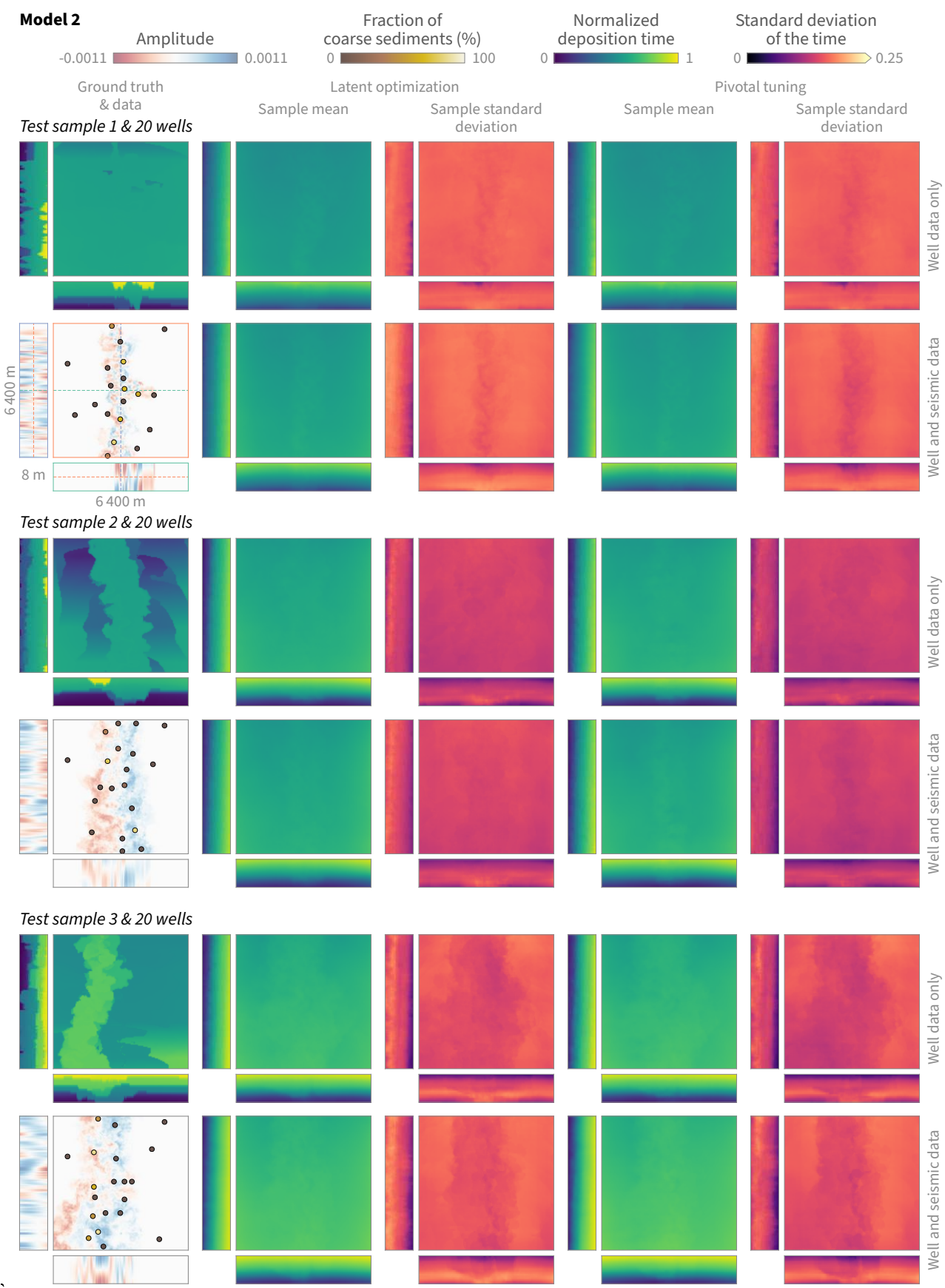

**Figure E.7** Comparison between the mean and standard deviation of the normalized deposition time for 300 samples inverted from the GAN model 2 with 20 wells for two inversion approaches without and with seismic data. The slices go through the center of the samples, as shown on the data with 4 wells.

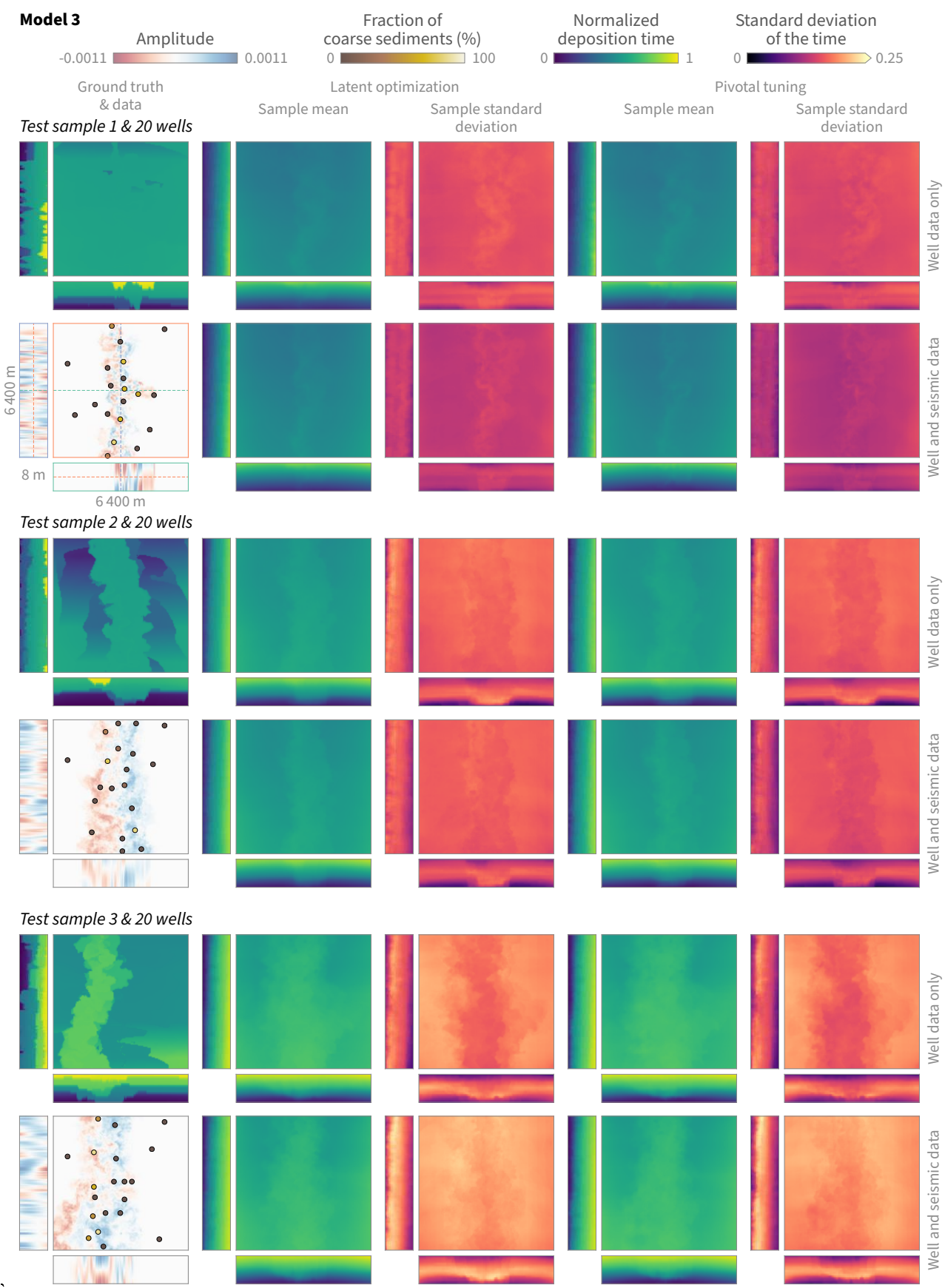

**Figure E.8**  Comparison between the mean and standard deviation of the normalized deposition time for 300 samples inverted from the GAN model 3 with 20 wells for two inversion approaches without and with seismic data. The slices go through the center of the samples, as shown on the data with 4 wells.

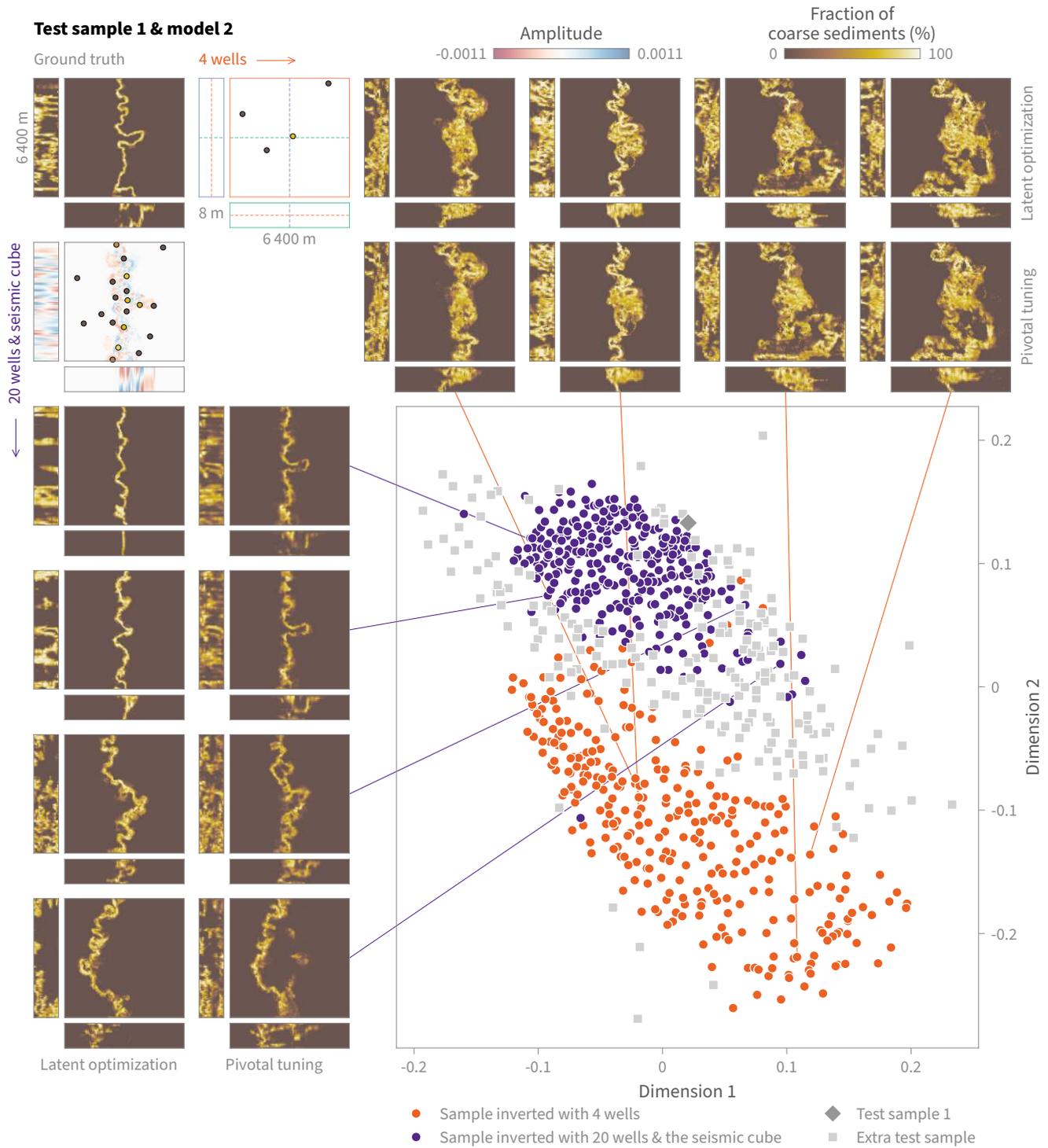

**Figure E.9** Testing sample quality and diversity from model 2 using mutlidimensional scaling to represent the sliced Wasserstein distances between 200 test samples and 300 samples inverted using pivotal tuning based on 4 wells or 20 wells and the seismic cube from test sample 1. The samples shown were randomly selected. The slices go through the center of the samples, as shown on the data with 4 wells.

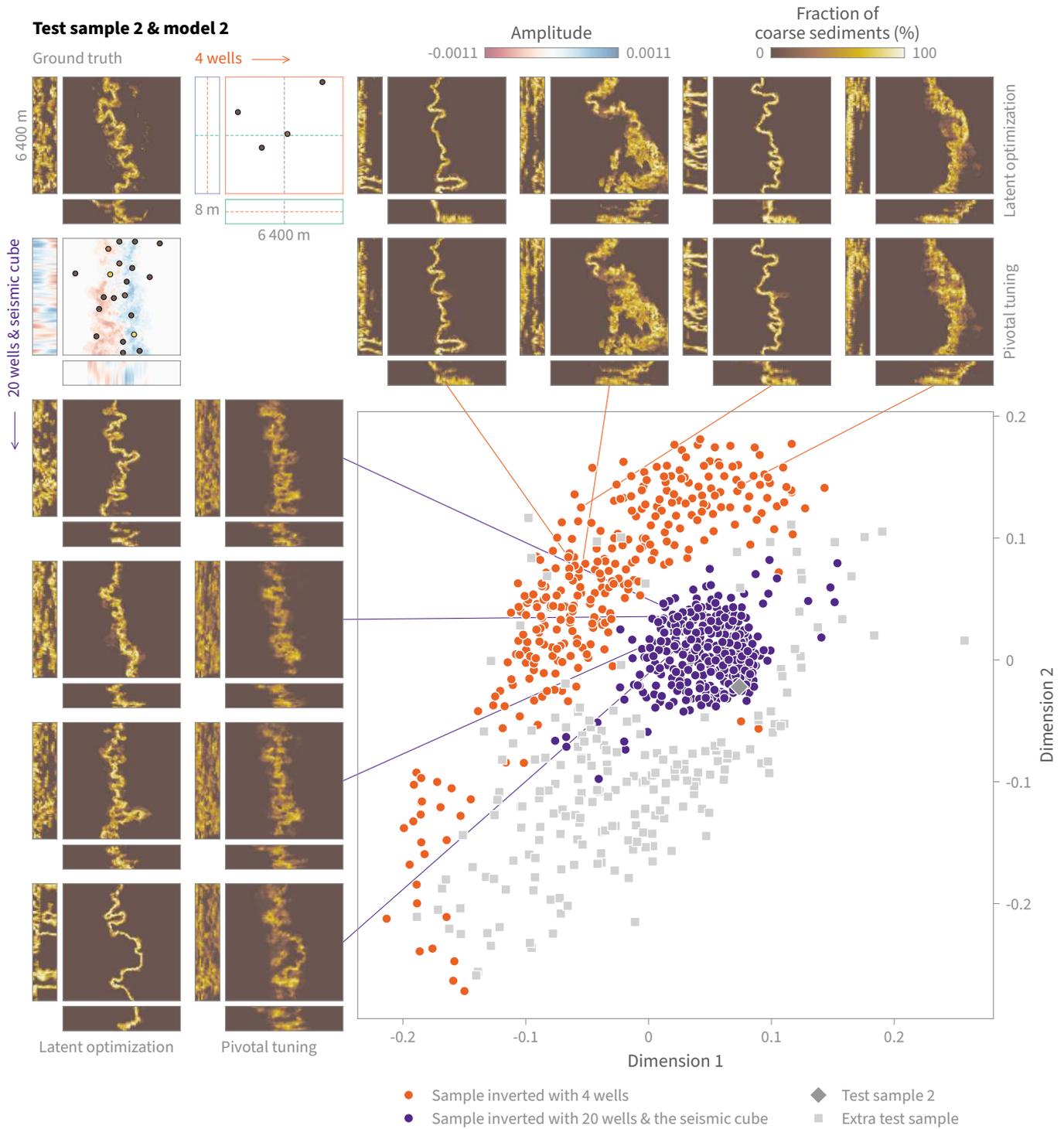

**Figure E.10** Testing sample quality and diversity from model 2 using mutlidimensional scaling to represent the sliced Wasserstein distances between 200 test samples and 300 samples inverted using pivotal tuning based on 4 wells or 20 wells and the seismic cube from test sample 2. The samples shown were randomly selected. The slices go through the center of the samples, as shown on the data with 4 wells.

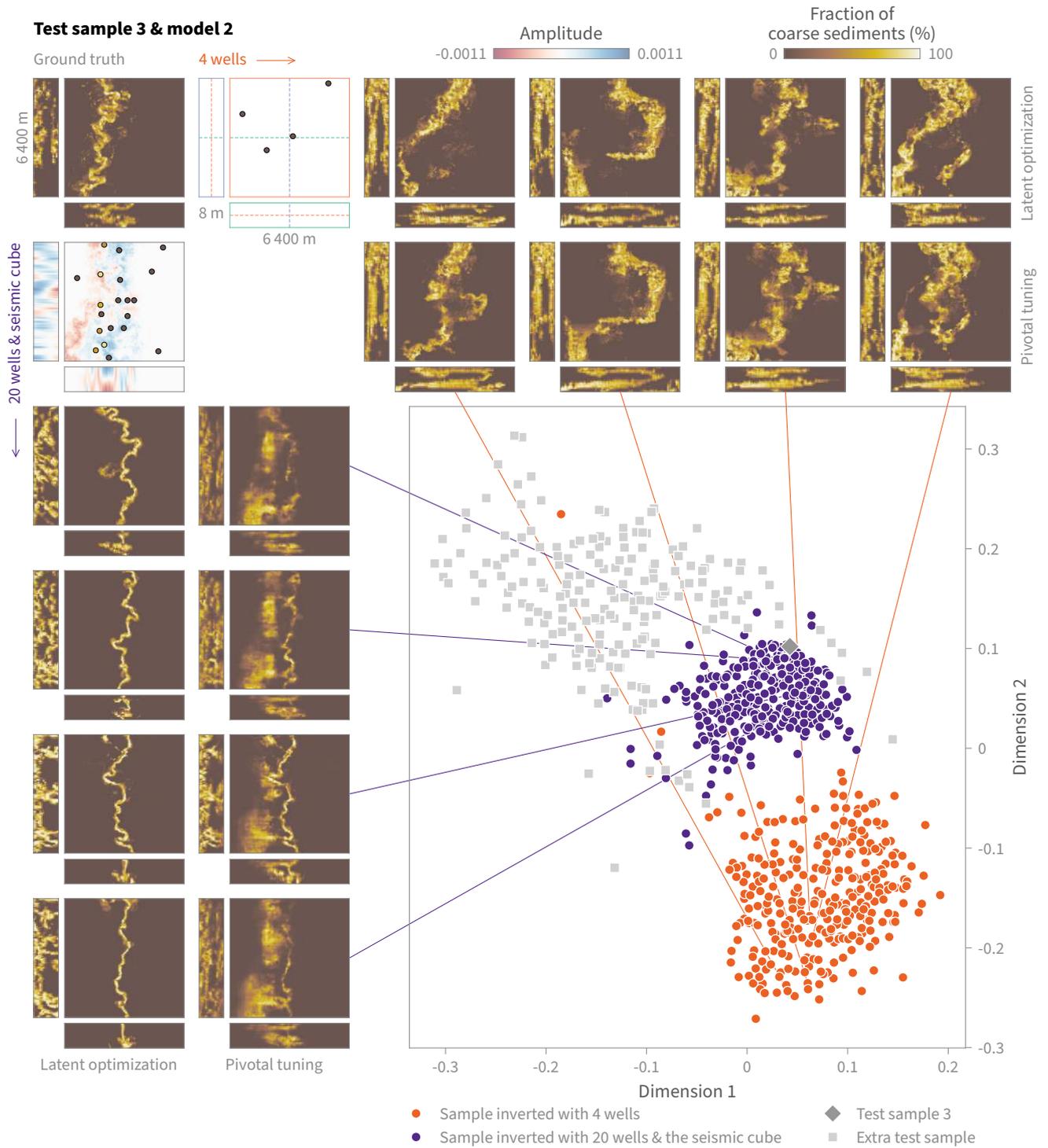

**Figure E.11** Testing sample quality and diversity from model 2 using mutlidimensional scaling to represent the sliced Wasserstein distances between 200 test samples and 300 samples inverted using pivotal tuning based on 4 wells or 20 wells and the seismic cube from test sample 3. The samples shown were randomly selected. The slices go through the center of the samples, as shown on the data with 4 wells.

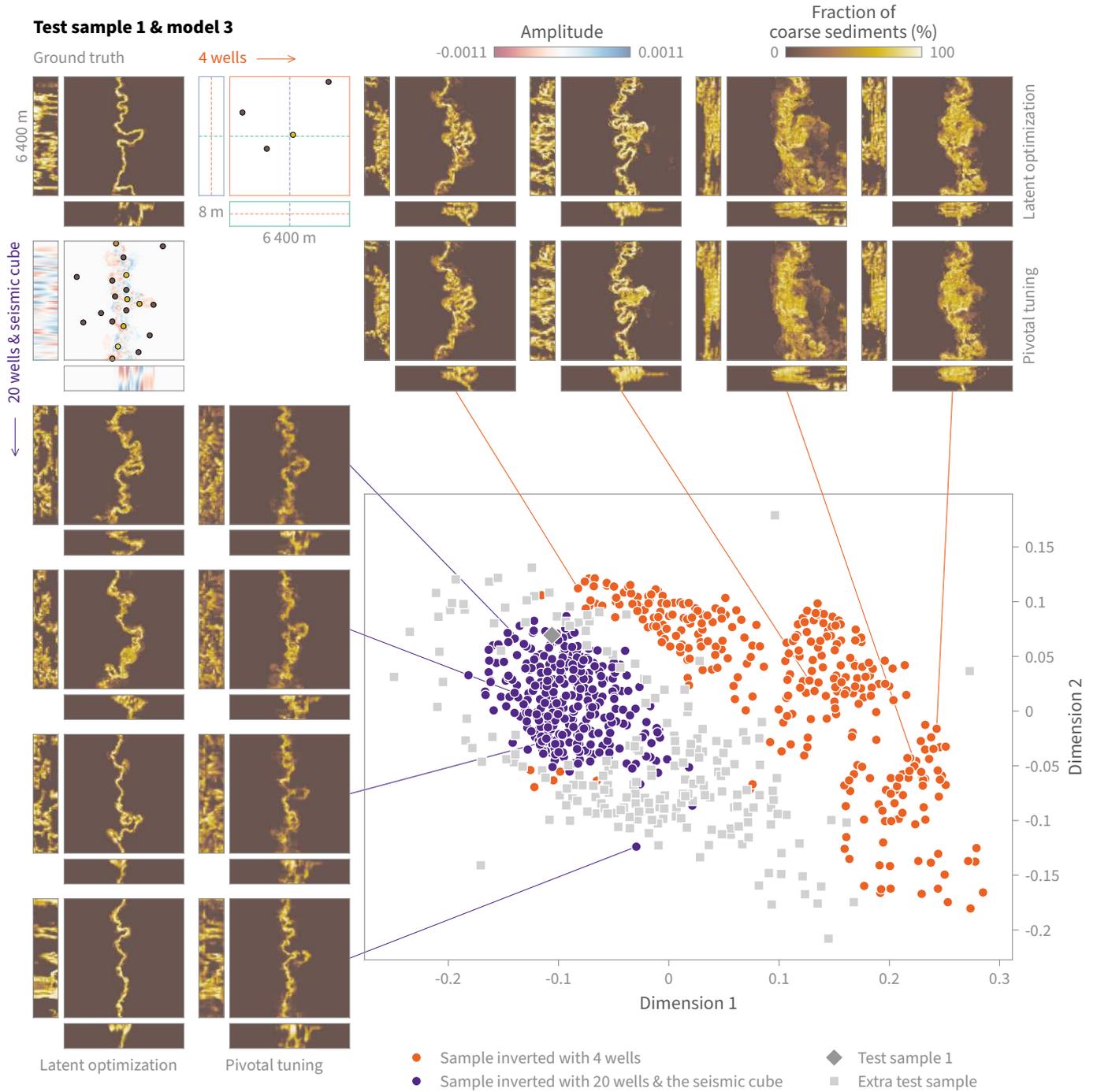

**Figure E.12** Testing sample quality and diversity from model 3 using mutlidimensional scaling to represent the sliced Wasserstein distances between 200 test samples and 300 samples inverted using pivotal tuning based on 4 wells or 20 wells and the seismic cube from test sample 1. The samples shown were randomly selected. The slices go through the center of the samples, as shown on the data with 4 wells.

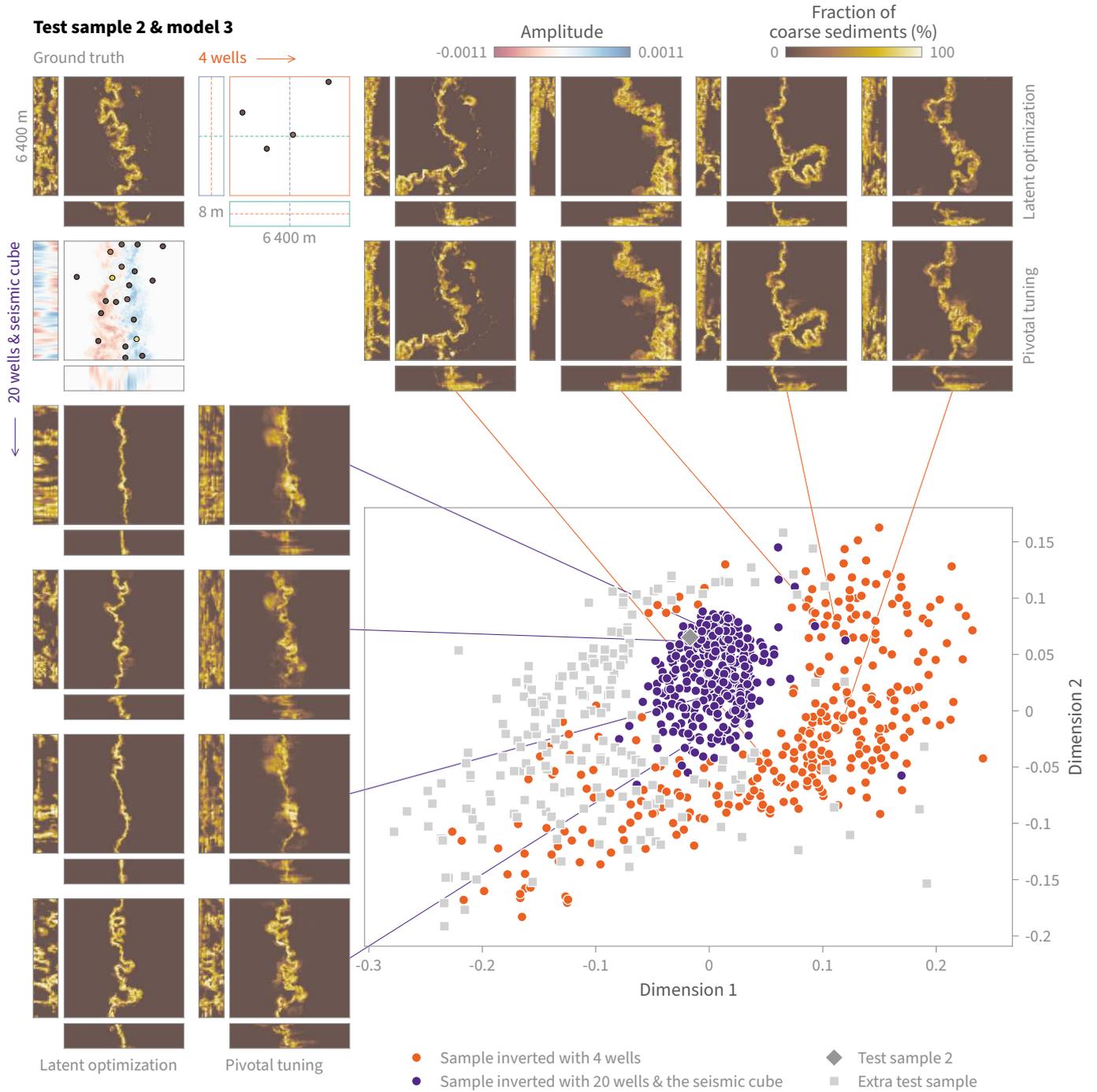

**Figure E.13** Testing sample quality and diversity from model 3 using mutlidimensional scaling to represent the sliced Wasserstein distances between 200 test samples and 300 samples inverted using pivotal tuning based on 4 wells or 20 wells and the seismic cube from test sample 2. The samples shown were randomly selected. The slices go through the center of the samples, as shown on the data with 4 wells.

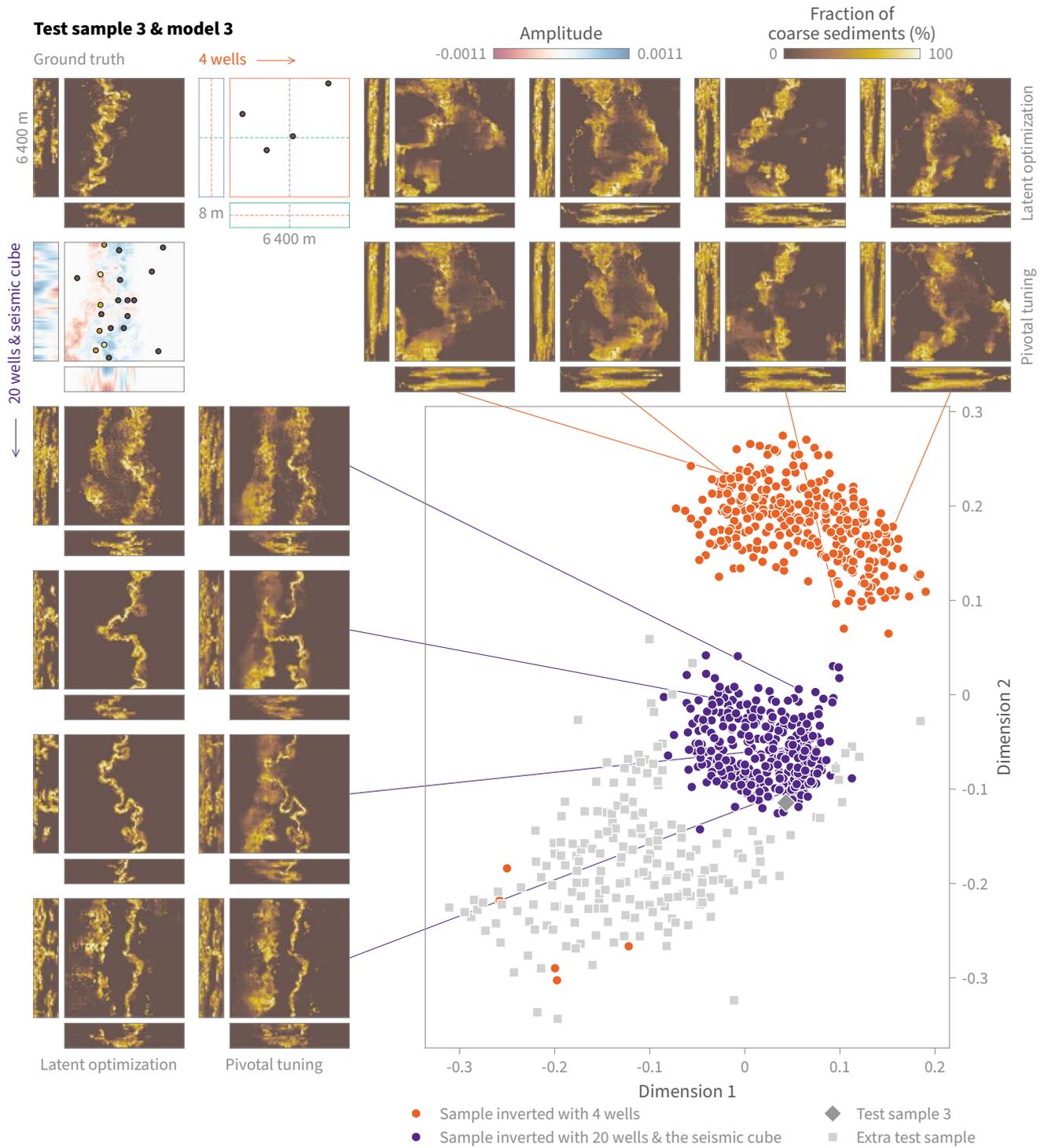

**Figure E.14** Testing sample quality and diversity from model 3 using mutlidimensional scaling to represent the sliced Wasserstein distances between 200 test samples and 300 samples inverted using pivotal tuning based on 4 wells or 20 wells and the seismic cube from test sample 3. The samples shown were randomly selected. The slices go through the center of the samples, as shown on the data with 4 wells.